\documentclass{article}

\PassOptionsToPackage{numbers, compress}{natbib}

 \usepackage[main, final]{neurips_2026}

\usepackage[utf8]{inputenc} 
\usepackage[T1]{fontenc}    
\usepackage{hyperref}       
\usepackage{url}            
\usepackage{booktabs}       
\usepackage{amsfonts}       
\usepackage{nicefrac}       
\usepackage{microtype}      
\usepackage{xcolor}         

\usepackage{multicol}
\usepackage{multirow}
\usepackage{siunitx}
\usepackage{graphicx}
\usepackage{subcaption}
\usepackage{tcolorbox}
\usepackage{enumitem}
\usepackage{titletoc}

\usepackage{subcaption} 
\usepackage{graphicx}
\usepackage{amsmath} 
\usepackage{colortbl}

\newcommand{\std}[1]{\ensuremath{{\scriptscriptstyle \pm #1}}}
\newcommand{\pmsd}{\ensuremath{\pm}}

\newcommand{\tv}{\boldsymbol{\tau}}

\newcommand{\revision}[1]{{\color{black}#1}}

\usepackage{xcolor}
\newcommand{\pos}{\color{green!60!black}}

\title{Learning Cultural Vectors for Cross-Cultural Generation}

\author{
Sina Malakouti \\
University of Pittsburgh \\
\texttt{sem238@pitt.edu}
\And
Deepti Ghadiyaram \\
Boston University \\
\texttt{dghadiya@bu.edu}
\AND
Boqing Gong \\
Boston University \\
\texttt{bgong@bu.edu}
\And
Adriana Kovashka \\
University of Pittsburgh \\
\texttt{kovashka@cs.pitt.edu}
}

\begin{document}

\maketitle
\vspace{-30pt}
\begin{center}
\url{https://sinamalakouti.github.io/CulturalVectors}
\end{center}
\vspace{30pt}

\begin{abstract}
Text-to-image diffusion models often struggle to faithfully represent diverse cultural contexts. We study culturally grounded image generation through cultural vectors, defined as the difference between model parameters fine-tuned on culturally grounded data and their pre-trained counterparts. First, we show that scaling cultural vectors at inference time controls the trade-off between cultural alignment, exaggeration, and hallucination. This enables learning cultural vectors from synthetic data while mitigating the exaggerated cultural depictions that may arise from biases in the synthetic images. Second, our analysis reveals that cultural knowledge is primarily encoded in deeper U-Net blocks, while activation-space analysis reveals a strong shared structure across countries. Finally, we investigate the compositionality of cultural vectors. We find that unlike prior work on task arithmetic, naively combining cultural vectors fails to produce meaningful multi-cultural behavior due to strong interference across cultures and degrades generalization to unseen countries. To reduce cultural interference, we introduce Cultural Activation Preservation (CAP), which encourages the preservation of other cultures' representations during training and improves compositionality. To improve generalization to unseen countries, we further introduce Descriptor-Guided Merging (DGM), a culturally aware approach that leverages textual descriptors to estimate cross-cultural similarities and merge cultural vectors. DGM improves both Task Arithmetic and TIES, with TIES+DGM achieving the lowest mean rank in human evaluation and ranking first in 10 of 11 countries.
\end{abstract}

\section{Introduction}
\label{sec:intro}
Consider the depiction of a simple act of a handshake in Iran. \revision{A text-to-image (T2I) model may place a mosque in the background, even when the prompt does not specify a religious setting.} 
Although recent T2I models~\cite{nano-banana,gpt-image-1,dalle3,flux2024,podell2023sdxl,esser2024scaling,rombach2022high} have fundamentally redefined the boundaries of photorealism and creativity, they remain \textbf{unreliable} when generating culturally grounded content. Models frequently depict incorrect actions or objects, stereotypical traditional attire, exaggerated scenes, and religious elements unrelated to the scene and activity. 
Recent works have primarily focused on benchmarking and evaluating cultural biases in T2I models, revealing systematic limitations in representing diverse cultural concepts, particularly for underrepresented cultures~\cite{basu2025geodiv,kannen_beyond_nodate,rege_cure_2025,nayak_culturalframes_2025,malakouti2026culture}. \revision{These findings motivate a broader investigation into how cultural information can be learned and controlled within T2I generative models.} 

In this work, we ask \revision{whether diffusion models can learn meaningful \textit{cultural representations}}, and if so, can it be \emph{controlled} to improve cultural faithfulness. Answering these questions is important for developing principled approaches to improve cultural faithfulness across diverse cultures while mitigating the amplification of stereotypes in generated images.
However, addressing these questions is challenging for two key reasons. First, cultural attributes are expressed through multiple interacting visual elements, such as clothing, objects, and scenes, making them difficult to control during generation. Second, high-quality culturally grounded data is \textbf{scarce}, \revision{particularly for underrepresented regions, while synthetic alternatives may inherit biases from their source models}.

To investigate these questions, \revision{we conduct a systematic study}, modeling cultural representation as parameter differences between fine-tuned and pretrained models. We refer to these as \textit{cultural vectors}. 
Inspired by task arithmetic~\cite{ilharco2023editing}, this formulation provides two potential advantages that directly address the challenges above. First, cultural vectors can be scaled at inference time to control the strength of cultural adaptation. 
Second, vectors learned for different countries can be combined to investigate multi-cultural generation and generalization to countries without target-specific training data.
However, while task vectors enable editing and composition in foundation models, this task (cultural generation) is \emph{unique in two ways}: (1) we are interested in generation (via diffusion models), and (2) capturing cultural knowledge in settings where high-quality visual training data is limited, despite the broader availability of cultural knowledge. These challenges motivate a systematic investigation into how cultural representations can be learned, controlled, and composed within diffusion models.

Towards this goal, we first investigate the \textbf{properties} of cultural vectors along multiple dimensions: (1) we assess their \textbf{validity} by examining whether their similarities reflect cultural relationships; (2) we investigate the \textbf{sources} of data for learning them (real or synthetic); (3) we study how inference-time \textbf{scaling} affects cultural alignment, exaggeration, and hallucination; and (4) we analyze their \textbf{structure} across U-Net layers and in parameter and activation spaces to understand how cultural information is represented. \revision{Second, we study \textbf{compositionality}, investigating whether combined cultural vectors can exhibit multi-cultural behavior and generalize to unseen countries. We find that naive merging leads to cross-cultural interference, motivating two complementary approaches: (1) We propose \textit{Cultural Activation Preservation} (CAP), which regularizes culture-specific fine-tuning to preserve the pretrained model's internal representations of cultures beyond the target culture, thereby reducing interference during vector composition. (2) We further develop \textit{Descriptor-Guided Merging} (DGM), which leverages textual descriptors to estimate cross-cultural similarities and guide vector composition for unseen countries.}

We conduct our analysis and compare the proposed methods on CULTIVate~\cite{malakouti2026culture} and CulturalFrames~\cite{nayak_culturalframes_2025} benchmarks.
Our analysis reveals several key findings: 
(1) Cultural vectors enable controllable generation, improving cultural alignment while mitigating exaggeration. 
(2) Synthetic data provides an effective signal to learn these vectors, especially when high quality real data is unavailable.
\revision{(3) Cultural vectors reflect meaningful cultural similarities, with stronger cultural information concentrated in specific deeper U-Net blocks. Moreover, 
cultural vectors exhibit a shared structure in activation space.}
(4) \revision{Naive merging suffers from cross-cultural interference, reducing faithfulness to individual cultures, while existing merging methods underperform the pretrained model when generalizing to unseen countries. CAP reduces interference, whereas DGM improves cross-cultural generalization, highlighting the importance of culturally aware composition. Human evaluation further supports DGM, with TIES+DGM achieving the lowest mean rank and ranking first in 10 of 11 countries.}


In summary, our contributions are threefold:
(1) We introduce \textit{cultural vectors} as a framework for studying cultural representations in diffusion models, showing that they can be learned from synthetic data and controlled through inference-time scaling to improve cultural faithfulness while mitigating exaggeration.
(2) We systematically investigate how cultural information is encoded within diffusion models, analyzing cultural similarities, block-wise representations, and the structure of cultural vectors in parameter and activation spaces.
(3) \revision{We study the compositionality of cultural vectors, revealing cross-cultural interference under naive merging. We introduce CAP and DGM to improve multi-cultural composition and generalization to unseen countries, respectively.}
\section{Related Works}
\label{sec:related_works}

\textbf{Cultural bias in T2I models.}
Recent works~\cite{malakouti2026culture,nayak_culturalframes_2025,rege_cure_2025,kannen_beyond_nodate,basu2025geodiv,jha_visage_2024,khanuja_image_2024} have studied cultural bias in T2I models, showing that current models suffer from limited cultural diversity~\cite{kannen_beyond_nodate,basu2025geodiv} and fail to faithfully generate culturally grounded concepts~\cite{rege_cure_2025,khanuja_image_2024,nayak_culturalframes_2025,malakouti2026culture}, particularly for underrepresented regions. 
Specifically,~\cite{nayak_culturalframes_2025,malakouti2026culture} show that models struggle with cultural scenarios that require context (e.g., social activities and social norms), often producing plausible yet culturally incorrect or exaggerated images. 
However, existing work primarily focuses on evaluating these failures through benchmarks and metrics, with limited understanding of how cultural knowledge is represented and controlled within diffusion models. 
In this work, we study cultural representations directly in model parameter space through cultural vectors, enabling controllable cultural generation and cross-cultural generalization.

\textbf{Debiasing diffusion models.} Various methods have been proposed to mitigate harmful or biased generation in diffusion models, primarily focusing on gender, racial, and safety biases~\cite{Schramowski_2023_CVPR,Jiang_2025_ICCV,Li_2024_CVPR}. 
In contrast, cultural generation remains heavily underexplored and is more challenging due to its context-dependent and compositional nature. 
Concretely,~\cite{Liu_2024_CVPR} improve cultural generation through contrastive finetuning on culturally grounded prompts, while~\cite{jeong-etal-2025-culture} enhances generation through culturally-aware prompt refinement using external cultural knowledge bases. However, these approaches rely on additional training data or external retrieval systems. Differently, we study how cultural knowledge is represented directly in model parameter space through cultural vectors, enabling controllable cultural generation and cross-cultural generalization.

\textbf{Weight interpolation and task arithmetic.} 
Recent work has shown that independently finetuned models often lie in connected regions of parameter space, enabling interpolation and merging across models without significant performance degradation~\cite{wortsman2022model,buzzega2025rethinking,choshen2022fusing,rame2401warm}. Building on this observation, task arithmetic~\cite{ilharco2023editing} represents task vectors as parameter differences between finetuned and pretrained models, enabling controllable editing, transfer, and multi-task composition through simple vector operations. A body of works~\cite{ortiz2023task,yadav2023tiesmerging,yu2024language,stoica2025knots} further proposed methods for addressing task interference and improving multi-task performance. 
However, application of task arithmetic and model merging remains understudied in T2I generative models.
Diffusion Soup~\cite{biggs2024diffusion} extends~\cite{wortsman2022model} to diffusion models, while Deme~\cite{ma2025decouple} learns and merges timestep-specific representations for improved diffusion finetuning on the same generation task, and~\cite{pham2024robust} utilizes task vectors for concept unlearning in diffusion models. In contrast, we extend task arithmetic~\cite{ilharco2023editing} to culturally grounded generation and study whether cultural knowledge can be represented, controlled, and composed in diffusion. To the best of our knowledge, this is the first work to study cultural representations in diffusion model parameter space and their use for cross-cultural generation.

\section{Methodology}
\label{sec:method}

\subsection{Cultural Vectors}
\label{sec:learn_tv}

\textbf{Preliminaries}.
Latent diffusion models (LDMs), e.g., Stable Diffusion~\cite{rombach2022high}, generate images by iteratively denoising Gaussian noise over $T$ timesteps. During training, the forward process gradually corrupts a clean latent $x_0$ by adding Gaussian noise $\epsilon$, yielding a noisy sample $x_t$:
\begin{equation}
x_t = \sqrt{\bar{\alpha}_t}\,x_0 + \sqrt{1-\bar{\alpha}_t}\,\epsilon,
\end{equation}
where $\epsilon \sim \mathcal{N}(0,\mathbf{I})$ and $\bar{\alpha}_t$ controls the noise level at timestep $t$.

The reverse process aims to 
recover $x_0$ 
by iteratively modeling posterior $p_\theta(x_{t-1} \mid x_t, p)$. In practice, 
a U-Net $\epsilon_\theta(x_t,p,t)$ predicts the noise at timestep $t$, conditioned on a text prompt $p$. The model is trained using the denoising objective:
\begin{equation}
\label{eq:diffusion_loss}
\mathcal{L}_{\mathrm{LDM}} =
\mathbb{E}_{t,x_0,p,\epsilon}
\left[
\left\|
\epsilon - \epsilon_\theta(x_t,p,t)
\right\|_2^2
\right].
\end{equation}

\begin{figure*}[t]
    \centering
    \includegraphics[width=0.9\textwidth]{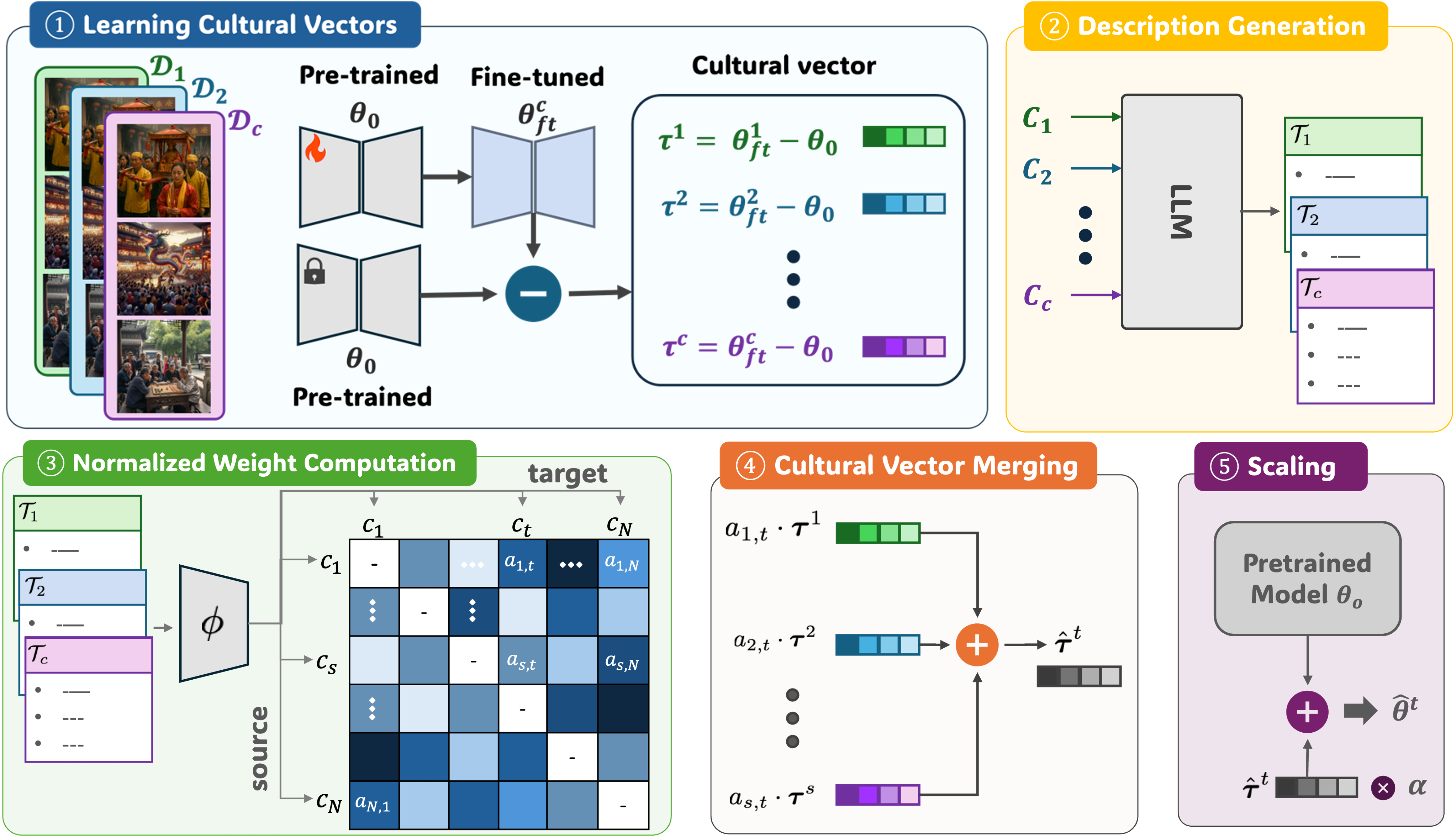}
    \caption{
\textbf{Overview.}
(1) For each country, pre-trained diffusion model is fine-tuned on cultural data to learn cultural vectors. (2) We generate cultural descriptors for each country using an LLM. (3) Descriptor embeddings are used to compute similarities between source and target countries. (4) Descriptor-Guided Merging (DGM) estimates a target cultural vector by weighted merging of culturally similar source vectors. (5) The resulting vector is scaled and applied to the pre-trained model at inference time to control generation.}
    \label{fig:method}
\end{figure*}

\textbf{Learning cultural vectors}. 
We investigate how cultural knowledge is encoded in diffusion models and whether it can be controlled to generate culturally faithful images.
Inspired by task arithmetic~\cite{ilharco2023editing}, we define \textit{cultural vectors} as the differences between culture-specific fine-tuned and pretrained model parameters. Let $\theta_0$ denote the pretrained U-Net parameters. For each country $c$, we construct a dataset $\mathcal{D}_c$ of culturally grounded image-caption pairs and fine-tune the model using Eq.~\ref{eq:diffusion_loss}, obtaining culture-specific parameters $\theta^c_{ft}$. The corresponding cultural vector is defined as:
\begin{equation}
    \tv^{c} = \theta^{c}_{ft} - \theta_{0}.
\end{equation}
At inference time, we apply the cultural vector to the pretrained model using a scaling coefficient $\alpha$:
\begin{equation}
    \hat{\theta}^{c} = \theta_{0} + \alpha \cdot \tv^{c}.
\end{equation}
The coefficient $\alpha$ controls the strength of cultural adaptation, enabling interpolation between the pretrained and the fine-tuned parameters without additional training. Throughout the paper we write $\tv = \theta_{ft} - \theta_{0}$ and apply $\theta_{0} + \alpha \cdot \tv$.

\subsection{Sources and Properties of Cultural Vectors}
\label{sec:method_propertis_tv}

Cultural vectors provide a mechanism for controlling cultural adaptation, but the data sources required to learn them and their properties remain to be understood. We study these properties along the following dimensions (results in Sec.~\ref{sec:res_properties}).

\textbf{Source of data.} 
For the dataset $\mathcal{D}_c$ in Sec.~\ref{sec:learn_tv}, one would most intuitively leverage real data samples from the respective countries. However, real images are difficult to collect, particularly for low-resource countries, and automated web crawling may result in some irrelevant samples being retrieved. 
Synthetic images provide an alternative but may inherit cultural biases and stereotypes from their source models. We compare both sources and analyze their effects on cultural alignment, exaggeration, and hallucination.

\textbf{Cultural similarity.}
To assess whether cultural vectors reflect meaningful cultural relationships, we measure the correlation between pairwise cultural vector similarities and cultural descriptor embedding similarities, using the latter as a proxy for cultural relatedness.

\textbf{Scaling and controllability.}
Next, we study how the scaling coefficient $\alpha$ affects generation behavior. By varying $\alpha$ at inference time, we analyze how the strength of the cultural representation influences cultural alignment, exaggeration, and hallucination.

\textbf{Block-wise cultural representations.}
To understand how cultural information is distributed across the U-Net, we learn cultural vectors from different blocks and compare their effect on cultural generation metrics from~\cite{malakouti2026culture} (alignment, exaggeration, and hallucination).

\textbf{Structure of cultural vectors \& activation vectors.} 
To understand the structure of cultural representations, we analyze cultural vectors in both parameter and activation spaces.
Following~\cite{ilharco2023editing}, we descriptively check whether cultural vectors concentrate in a shared low-dimensional subspace of parameter space. We perform SVD on $V \in \mathbb{R}^{C \times D}$, where each row is a flattened cultural vector $\tv^{c}$ and $D$ is the number of parameters, and compute the explained variance ratio and effective rank:
\begin{equation}
\mathrm{EVR}_k =
\frac{\sigma_k^2}{\sum_j \sigma_j^2},
\qquad
r_{\mathrm{eff}} =
\exp\left(-\sum_k q_k \log q_k\right),
\quad q_k = \mathrm{EVR}_k,
\end{equation}
where $\sigma_k$ denotes the $k$-th singular value.
\revision{Similarly, we obtain activation vectors for each U-Net block by averaging the differences between the activations of the culturally adapted model $\theta_0+\alpha\tv^c$ and the pretrained model $\theta_0$ over a shared set of probe samples. We stack the flattened activation vectors across countries into a matrix $A_l$, where $l$ denotes the U-Net block, and apply the same SVD analysis. Following~\cite{stoica2025knots}, we analyze CKA~\cite{kornblith2019similarity} in the appendix.}

\subsection{Compositionality of Cultural Vectors}
\label{sec:cc_gen}

We study whether cultural vectors can be composed to enable multi-cultural generation and generalization to unseen countries. Although prior work has shown that models obtained by linearly combining task vectors can exhibit multi-task behavior~\cite{ilharco2023editing}, composing cultural vectors may lead to cross-cultural interference. To improve compositionality, we explore two complementary approaches: preserving pretrained representations of non-target cultures during fine-tuning and leveraging cross-cultural similarities to guide vector merging. We present the results in Sec.~\ref{sec:result_cross_cultural_generation}.

\textbf{Composition of cultural vectors.}
Given a set of countries $\mathcal{C}$, we define the merged cultural vector as:
\begin{equation}
\tv_{\mathrm{merged}} =
\sum_{c \in \mathcal{C}} a_c \tv^c,
\end{equation}
where $a_c$ is the merging weight for country $c$.
By default, we use uniform averaging ($a_c = 1/|\mathcal{C}|$). We evaluate composition in two settings: (1) merging subsets of countries and evaluating faithfulness on both the merged countries and the full country pool, and (2) leave-one-country-out generalization, where source vectors are combined to estimate a vector for a held-out target country.

\revision{\textbf{Cultural Activation Preservation (CAP).}
We introduce CAP, an activation-preservation regularization that encourages adaptation to the target culture while preserving the pretrained model's internal representations for non-target cultures.
Specifically, when learning $\tv^c$ for country $c$, we additionally sample image-caption pairs $(x'_0,p')$ from countries $\mathcal{C}\setminus\{c\}$ and penalize deviations from the pretrained U-Net's activations:
\begin{equation}
\mathcal{L}_{\mathrm{CAP}} =
\mathbb{E}_{t,p',x'_0,\epsilon}
\left[
\frac{1}{L}\sum_{l=1}^{L}
\frac{
\left\|h_l^\theta(x'_t,p',t)
-h_l^{\theta_0}(x'_t,p',t)\right\|^2
}{
\left\|h_l^{\theta_0}(x'_t,p',t)\right\|^2
}
\right].
\end{equation}
$h_l^\theta$ and $h_l^{\theta_0}$ denote the fine-tuned and frozen pretrained U-Net activations at block $l$, respectively, over all $L$ down, mid, and up blocks. The final objective is defined as: 
\begin{equation}
\label{eq:loss_total}
\mathcal{L}_{\mathrm{total}}
=
\mathcal{L}_{\mathrm{LDM}}
+
\lambda_{\mathrm{CAP}}\mathcal{L}_{\mathrm{CAP}},
\end{equation}
where $\lambda_{\mathrm{CAP}}$ controls regularization strength. This encourages each vector to learn its target culture while constraining changes on other countries.}

\textbf{Descriptor-Guided Merging (DGM).}
We investigate whether cultural similarity can guide composition when no cultural vector is available for a target country $t$. Given source countries $\mathcal{S}=\mathcal{C}\setminus\{t\}$, we use textual descriptors of cultural attributes, including attire, objects, actions, and scenes, following~\cite{malakouti2026culture}. For each country $c$, we average its descriptor embeddings:
\begin{equation}
    \mathbf{z}_c = \frac{1}{|\mathcal{T}_c|}
    \sum_{d \in \mathcal{T}_c} \phi(d),
\end{equation}
where $\mathcal{T}_c$ is the descriptor set and $\phi$ is a text encoder. We use normalized, positive mean-centered similarities between the target and source representations as merging weights:
\begin{equation}
\tilde{a}_{s,t} =
\max\big(0,\operatorname{sim}(\mathbf{z}_s,\mathbf{z}_t)-\mu_t\big),
\qquad
a_{s,t} = \frac{\tilde{a}_{s,t}}
{\sum_{s'\in\mathcal{S}}\tilde{a}_{s',t}}.
\end{equation}
Here, $\mu_t=\frac{1}{|\mathcal{S}|}
\sum_{s\in\mathcal{S}}
\operatorname{sim}(\mathbf{z}_s,\mathbf{z}_t)$
is the mean source similarity. The estimated target vector is
\begin{equation}
    \hat{\tv}^t = \sum_{s\in\mathcal{S}} a_{s,t}\cdot\tv^s.
\end{equation}
Descriptors determine only the merging weights; DGM requires neither target-country training images nor additional fine-tuning.

\section{Experimental Setup}
\label{sec:exp_setup}

\subsection{Datasets}

\textbf{CULTIVate}~\cite{malakouti2026culture} covers 576 activities across 9 categories and 16 countries, comprising 5 Global North and 11 Global South countries. Each activity includes culturally grounded descriptors and real images. The benchmark includes 6 T2I models: SD3-Medium~\cite{esser2024scaling}, FLUX~\cite{flux2024}, Qwen-Image~\cite{wu2025qwen}, DALL-E 3~\cite{dalle3}, GPT-Image-1~\cite{gpt-image-1}, and Gemini 2.5 Flash~\cite{nano-banana}. We utilized synthetic and real images as training data.

\textbf{CulturalFrames} ~\cite{nayak_culturalframes_2025} covers 10 countries (Iran, China, Japan, South Africa, Brazil, Chile, Canada, Poland, Germany).
 CulturalFrames includes 983 prompts in family, religion, and dates of significance categories, paired with 3,637 images generated by three T2I models (Flux~\cite{flux2024}, SD3-Large~\cite{esser2024scaling}, and Imagen3~\cite{baldridge2024imagen}).
These image-text pairs are utilized as our training dataset $\mathcal{D}$ to learn cultural vectors.

\subsection{Implementation Details}
The models utilize Stable Diffusion v1.5~\cite{rombach2022high} as backbone, generating images at $512\times512$ resolution with 30 denoising steps and a classifier-free guidance (CFG) scale~\cite{ho2021classifier} of 7.5. Cultural vectors are learned from U-Net parameters, while the VAE and text encoder remain frozen. We fine-tune the U-Net for 1,500 iterations using a learning rate of $10^{-5}$, batch size of 4, and mixed precision on a single NVIDIA L40S GPU per country. Inference-time scaling and vector merging require
no additional training. For CAP, we set the regularization coefficient $\lambda_{\mathrm{CAP}}=0.1$. For CULTIVate~\cite{malakouti2026culture}, we use 100 synthetic images per culture generated by three proprietary models, and for CulturalFrames~\cite{nayak_culturalframes_2025}, we use 363 images per culture generated by three models. We additionally examine the effect of different image generators, including publicly available models, in the appendix. We use $\alpha=0.7$ for all methods and baselines to ensure a consistent comparison and study its effect in Tab.~\ref{tab:full_tv_alpha}. All evaluation images are generated with a fixed random seed of 42 for reproducibility. For TIES~\cite{yadav2023tiesmerging}, we follow the original setup, retaining the top $k=50\%$ of parameter changes and using  sum of values to determine the sign.

\subsection{Evaluation}
Evaluating cultural faithfulness is challenging due to its subjectivity and limited ground truth images. We use two complementary approaches. \textbf{Automatic evaluation} follows~\cite{malakouti2026culture} (best metric on this task and dataset), using GPT-4o~\cite{hurst2024gpt} to rate cultural alignment (ALIGN), exaggeration (EXAG), and hallucination (HAL) on a 1--5 Likert scale. We compute $\text{FAITH}=\frac{1}{3}(\text{ALIGN}+(6-\text{HAL})+(6-\text{EXAG}))$, averaging scores from two evaluations per image. \textbf{Human evaluation} uses native annotators recruited through Prolific in two studies: (1) ranking pretrained, TIES, and TIES+DGM generations across 301 prompts from 11 countries, and (2) comparing synthetic and real cultural vectors across 274 prompts from 10 countries. Images are presented in randomized order, with attention checks and at least two annotators per form. 


\section{Results}
\label{sec:results}
\begin{table}[tp]
\setlength{\tabcolsep}{3pt}
\renewcommand{\arraystretch}{1.2}
\caption{\textbf{Cultural vectors capture cultural knowledge across datasets.} We evaluate on CULTIVate and CulturalFrames. $\alpha=0.7$, $\uparrow$/$\downarrow$ indicate higher/lower is better and $\pm$ is standard deviation. For FAITH, we report relative improvement vs fine-tuning. CulturalFrames lacks real images.
}
\centering
\scriptsize
\begin{tabular}{ll||ccc|c||ccc|c}
\toprule
\multirow{2}{*}{\textbf{Method}} 
& \multirow{2}{*}{\textbf{Data}}
& \multicolumn{4}{c||}{\textbf{CULTIVate}}
& \multicolumn{4}{c}{\textbf{CulturalFrames}} \\
&
& ALIGN $\uparrow$ & EXAG $\downarrow$ & HAL $\downarrow$ & FAITH $\uparrow$
& ALIGN $\uparrow$ & EXAG $\downarrow$ & HAL $\downarrow$ & FAITH $\uparrow$ \\
\hline\hline

Pre-trained & --
& 3.44 \std{0.33} & 2.28 \std{0.31} & 1.92 \std{0.39} & 3.75
& 3.91 \std{0.15} & 1.75 \std{0.21} & 1.38 \std{0.15} & 4.26 \\

\hline\hline

\multirow{3}{*}{$\tv$}
& Random
& 3.48\std{0.33} & 2.27\std{0.35} &1.89\std{0.39} & 3.77
& 3.90 \std{0.12}& 1.75\std{0.20} & 1.70\std{0.14} & 4.15 \\

& Synth &
\textbf{4.56} \std{0.22} & 2.00 \std{0.26} & \textbf{1.18} \std{0.10} 
& \textbf{4.46} {\tiny{\pos (+4.2\%)}}
& 4.16 \std{0.11} & \textbf{1.73 \std{0.24}} & \textbf{1.25} \std{0.10} 
& \textbf{4.39} {\tiny{\pos (+0.9\%)}} \\

& Real
& 3.89 \std{0.25} & \textbf{1.94 \std{0.16}} & 1.57 \std{0.20} 
& 4.13 {\tiny{\pos(+7.3\%)}}
& N/A & N/A & N/A & N/A \\

\hline
\hline

\multirow{2}{*}{Fine-tune}
& Synth 
& 4.49 \std{0.25} & 2.39 \std{0.24} & 1.26 \std{0.17} & 4.28
& \textbf{4.19} \std{0.08} & 1.85 \std{0.25} & 1.28 \std{0.10} & 4.35 \\

& Real
& 3.73 \std{0.30} & 2.40 \std{0.22} & 1.77 \std{0.28} & 3.85
& N/A & N/A & N/A & N/A \\




\bottomrule
\end{tabular}
\label{tab:tv_poc}
\end{table}

\subsection{Can we learn meaningful cultural vectors? }
\label{sec:results_emergence}

\textbf{Cultural vectors capture cultural knowledge.}
Tab.~\ref{tab:tv_poc} shows that cultural vectors learned from real and synthetic data improve cultural faithfulness over the pretrained model and generally outperform full fine-tuning. Synthetic $\tv$ improves FAITH by 18.9\% (from 3.75 to 4.46) on CULTIVate and from 4.26 to 4.39 on CulturalFrames, outperforming both the pretrained and full fine-tuning baselines. These gains extend across Global North and Global South countries (Tab.~\ref{tab:full_tv_alpha} in Appendix), with synthetic $\tv$ matching or outperforming full fine-tuning in 15 of 16 countries (Tab.~\ref{tab:country_faithfulness} in Appendix) and achieving the highest FAITH across all nine activity categories (Tab.~\ref{tab:activity_faithfulness} in Appendix). In contrast, a random vector with matched mean and variance yields negligible or negative changes in FAITH ($+0.02$ on CULTIVate and $-0.11$ on CulturalFrames), suggesting that the improvements depend on the learned cultural vector rather than arbitrary parameter perturbations.

\begin{table*}[t]
\centering
\begin{minipage}[t]{0.4\textwidth}
\centering
\caption{\textbf{Generalization to new activities.}
Evaluation on 10 held-out activities per country in CULTIVate
($n=160$, $\alpha=0.7$).
$\uparrow$ and $\downarrow$ indicate that higher and lower values are better, respectively.}
\label{tab:direct_adaptation_unseen_activities}

\scriptsize
\setlength{\tabcolsep}{2pt}
\begin{tabular}{@{}lcccc@{}}
\toprule
Method & ALIGN $\uparrow$ & EXAG $\downarrow$
       & HAL $\downarrow$ & FAITH $\uparrow$ \\
\hline
\hline

Pre-trained       & 3.42 & \textbf{2.35} & 1.99 & 3.69 \\
$\tv$ (Synth)     & \textbf{3.70} & 2.44
                  & \textbf{1.79} & \textbf{3.82} \\
$\tv$ (Real)      & 3.51 & 2.42 & 2.05 & 3.68 \\
Fine-tune (Synth) & 3.62 & 2.78 & 1.88 & 3.65 \\
\bottomrule
\end{tabular}
\end{minipage}
\hfill
\begin{minipage}[t]{0.57\textwidth}
\vspace{0pt}
\centering
\includegraphics[width=\linewidth]{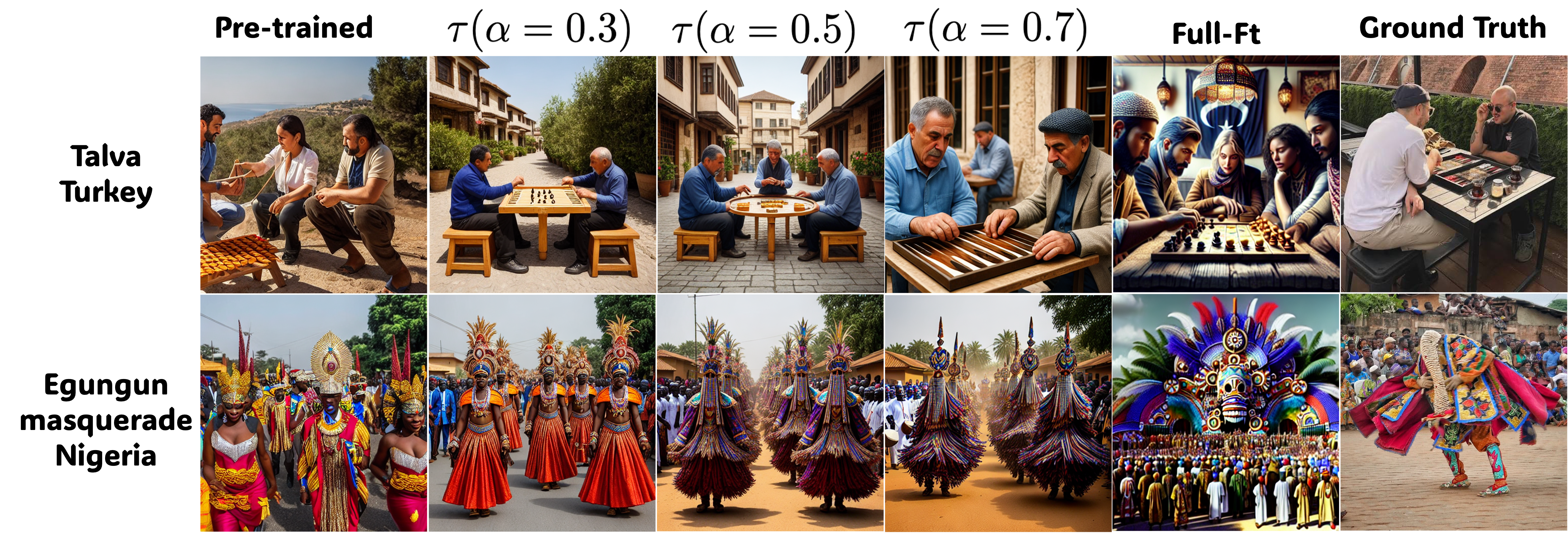}
\captionof{figure}{Inference-time scaling controls cultural alignment and exaggeration.}
\label{fig:qual_alpha_sweep}
\end{minipage}
\end{table*}


\revision{\textbf{Generalization to unseen activities.}
We further evaluate whether cultural vectors generalize to unseen activities. We split CULTIVate into train and test sets by randomly selecting 10 activities per country as held-out set. As shown in Tab.~\ref{tab:direct_adaptation_unseen_activities}, consistent with Tab.~\ref{tab:tv_poc} (in-domain), cultural vectors learned from synthetic data achieve the best performance, outperforming the pretrained model, full fine-tuning, and cultural vectors learned from real data.}

\textbf{Advantages synthetic vs. real training data.}
Tab.~\ref{tab:tv_poc} shows that cultural vectors learned from synthetic data outperform those learned from real data in alignment (ALIGN 4.56 vs. 3.89) and hallucination (HAL 1.18 vs. 1.57), while real $\tau$ achieves slightly lower exaggeration (EXAG 1.94 vs. 2.00). We further compare real and synthetic vectors through human evaluation (Tab.~\ref{tab:human_real_synthetic} in Appendix), where native annotators prefer images generated using synthetic $\tau$ in 62.1\% of head-to-head comparisons. These findings demonstrate the effectiveness of synthetic data for learning cultural vectors, while increasing exaggeration which can be controlled by scaling $\alpha$ (see Sec.~\ref{sec:res_properties}).

\subsection{Properties of cultural vectors}
\label{sec:res_properties}

\begin{figure}[t]
    \centering

    \begin{subfigure}[t]{0.32\linewidth}
        \centering
        \includegraphics[width=\linewidth]{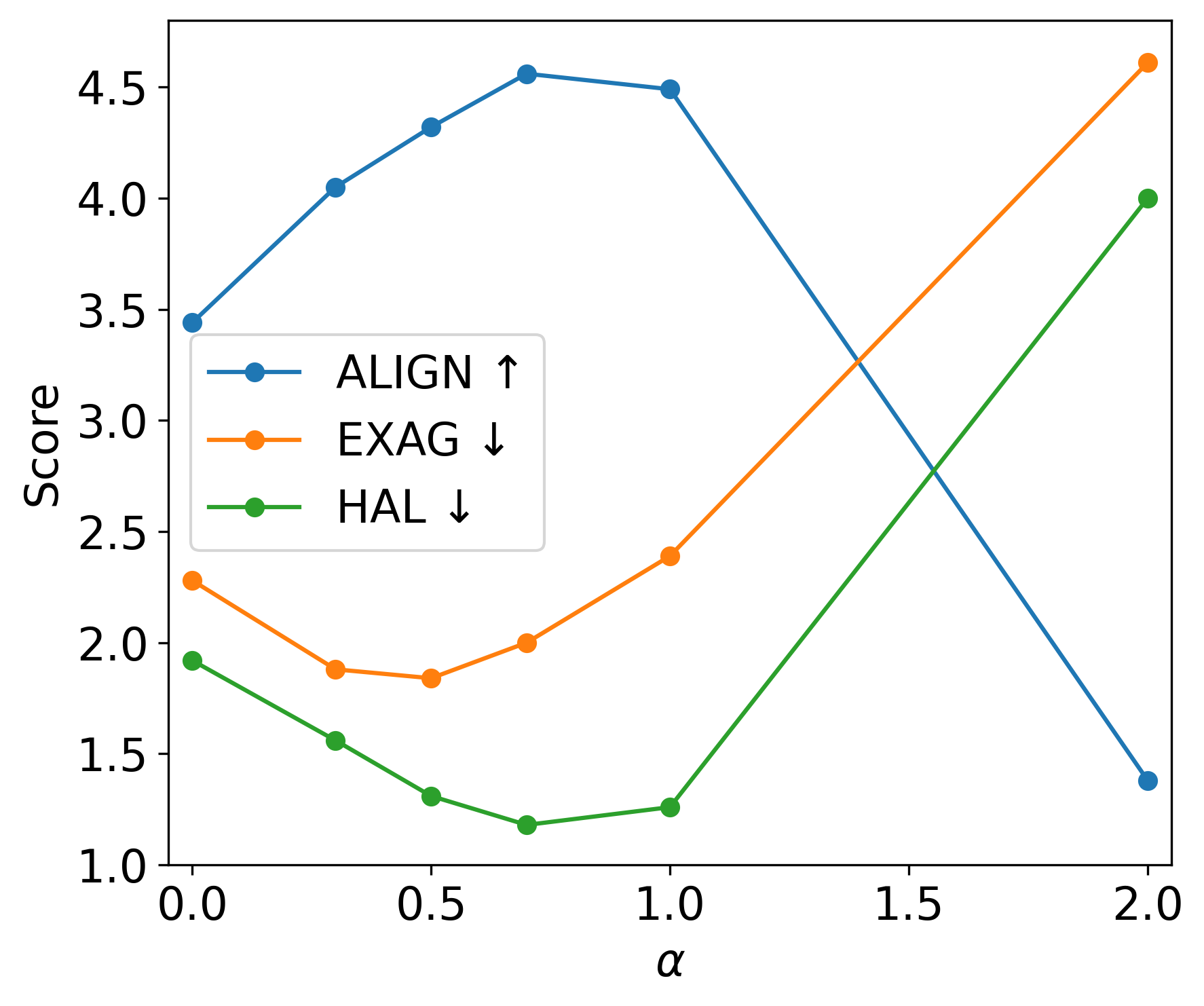}
        \caption{Ablation on $\alpha$.}
        \label{fig:alpha_ablation}
    \end{subfigure}
    \begin{subfigure}[t]{0.32\linewidth}
        \centering
        \includegraphics[width=0.8\linewidth]{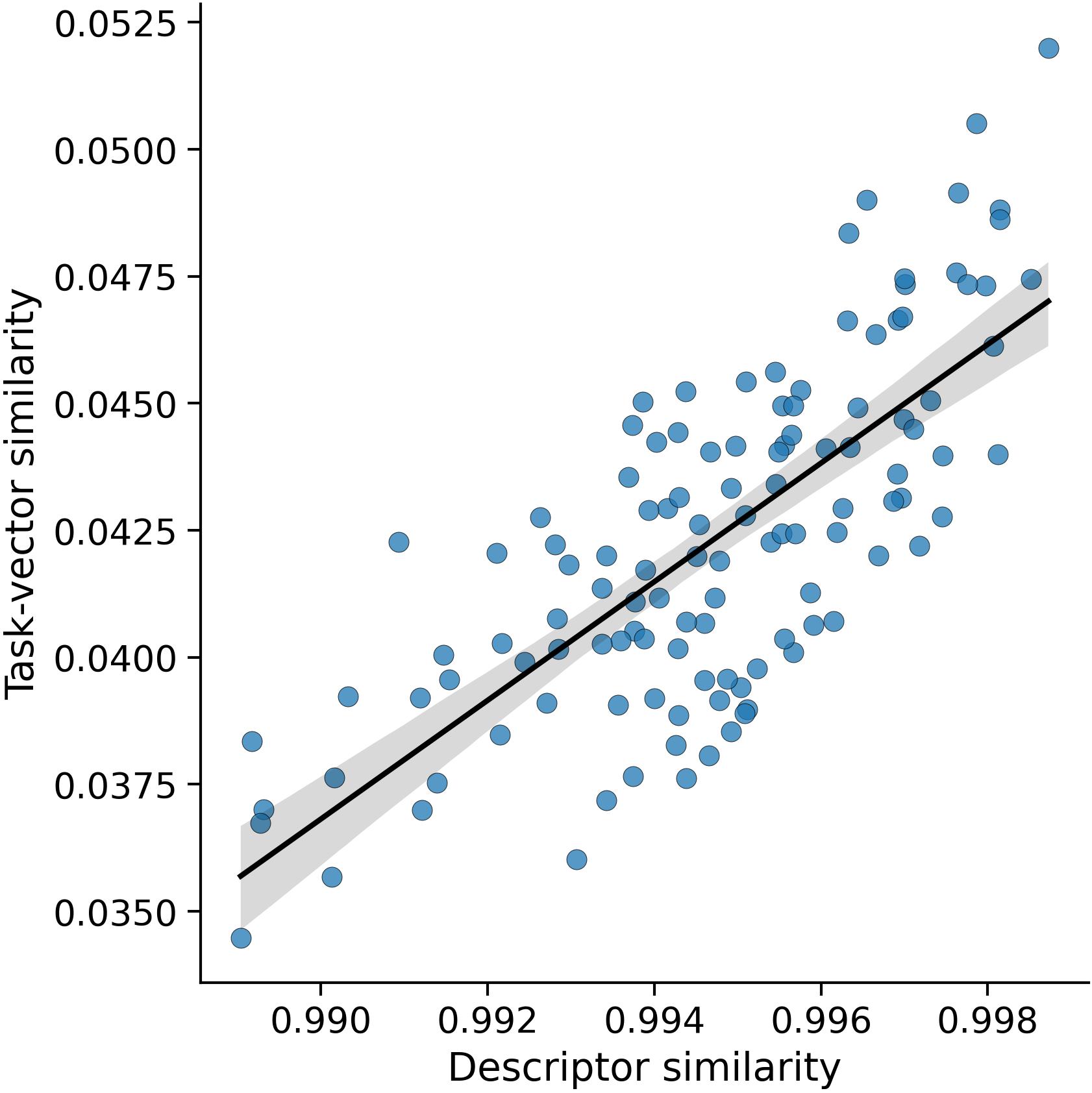}
        \caption{Pairwise similarity correlation.}
        \label{fig:similarity_corr}
    \end{subfigure}
    \begin{subfigure}[t]{0.32\linewidth}
        \centering
        \includegraphics[width=\linewidth]{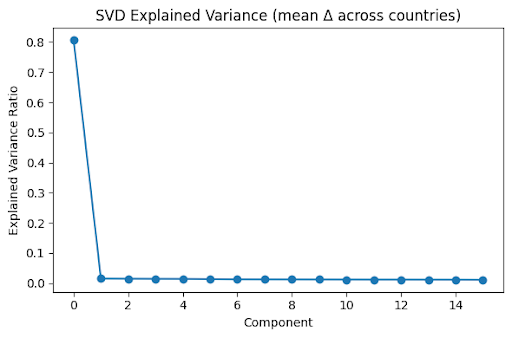}
        \caption{\revision{Activation-space spectrum.}}
        \label{fig:svd_activation}
    \end{subfigure}


    \caption{
\textbf{Properties of cultural vectors.}
(a) Scaling $\alpha$ controls cultural alignment, exaggeration, and hallucination.
(b) Vector similarities correlate strongly with cultural descriptor similarities.
\revision{(c) Activation shifts are dominated by a single principal direction.}}
    
    \label{fig:properties_cultural_vectors}
\end{figure}
\paragraph{Effect of $\alpha$ scaling.}
Fig.~\ref{fig:alpha_ablation} shows that scaling $\alpha$ enables gradual control over cultural adaptation. Cultural alignment and faithfulness generally improve up to $\alpha=0.7$, while larger values increase exaggeration and hallucination. \revision{Fig.~\ref{fig:qual_alpha_sweep} illustrates how increasing $\alpha$ can improve cultural relevance and introduce more appropriate cultural details than the pretrained model. However, excessive scaling (e.g., $\alpha=2$) can produce severely exaggerated and incoherent images (not shown).} 
Notably, scaling cultural vectors enables improving cultural faithfulness and exaggeration simultaneously.

\textbf{Cultural vectors encode meaningful cultural knowledge.} We measure pairwise cosine similarity between descriptor embeddings and cultural vectors across countries, using descriptor similarity (Sec.~\ref{sec:cc_gen}) as a proxy for cultural relatedness. Fig.~\ref{fig:similarity_corr} shows strong correlation between the two (Pearson $r=0.75$, Spearman $\rho=0.72$), confirming that cultural vectors capture culturally meaningful structure. Figs.~\ref{fig:heatmap_descriptors}/~\ref{fig:heatmap_tv} (Appendix) further support this, with culturally proximate countries clustering together in both similarity matrices.

\textbf{Shared activation-space structure across cultures.}
\revision{Fig.~\ref{fig:svd_activation} shows that activation vectors across countries exhibit a low-dimensional structure, with an effective rank (see in Sec.~\ref{sec:method_propertis_tv}) of 2.93 out of 16 possible directions. CAP increases the effective rank to 4.02, suggesting less concentrated activation shifts across cultures. These findings suggest a potential relationship between shared activation structure and cross-cultural interference. Detailed analysis is provided in Appendix Sec.~\ref{app:cultural_generalization}.}

\begin{figure}[t]
\centering

\begin{minipage}[t]{0.5\linewidth}
\vspace{0pt}
\centering
\scriptsize
\setlength{\tabcolsep}{4pt}

\captionof{table}{\textbf{Single-block training of cultural vectors.} Applying cultural vectors to different UNet blocks. Best result per up/down blocks \textbf{bolded}.}
\label{tab:tv_layerwise}

\begin{tabular}{lc||ccc}
\toprule
\textbf{Model}
& \textbf{Layer} 
& \textbf{HAL} $\downarrow$
& \textbf{EXAG} $\downarrow$
& \textbf{ALIGN} $\uparrow$ \\
\midrule
\multirow{9}{*}{$\tv$} 
& \texttt{down-block-0} & 1.93 & 2.38 & 3.54 \\
& \texttt{down-block-1} & 1.77 & 2.30 & 3.67 \\
& \texttt{down-block-2} & \textbf{1.49} & \textbf{2.06} & \textbf{4.13} \\
& \texttt{down-block-3} & 1.89 & 2.43 & 3.50 \\
\cline{2-5}
& \texttt{mid-block}    & 1.69 & 2.37 & 3.83 \\
\cline{2-5}
& \texttt{up-block-0}   & 1.87 & 2.41 & 3.65 \\
& \texttt{up-block-1}   & \textbf{1.28} & \textbf{1.88} & \textbf{4.39} \\
& \texttt{up-block-2}   & 1.74 & 2.13 & 3.76 \\
& \texttt{up-block-3}   & 1.88 & 2.24 & 3.54 \\
\bottomrule
\end{tabular}

\end{minipage}
\hfill
\begin{minipage}[t]{0.49\linewidth}
\vspace{0pt}
\centering

\includegraphics[width=\linewidth]{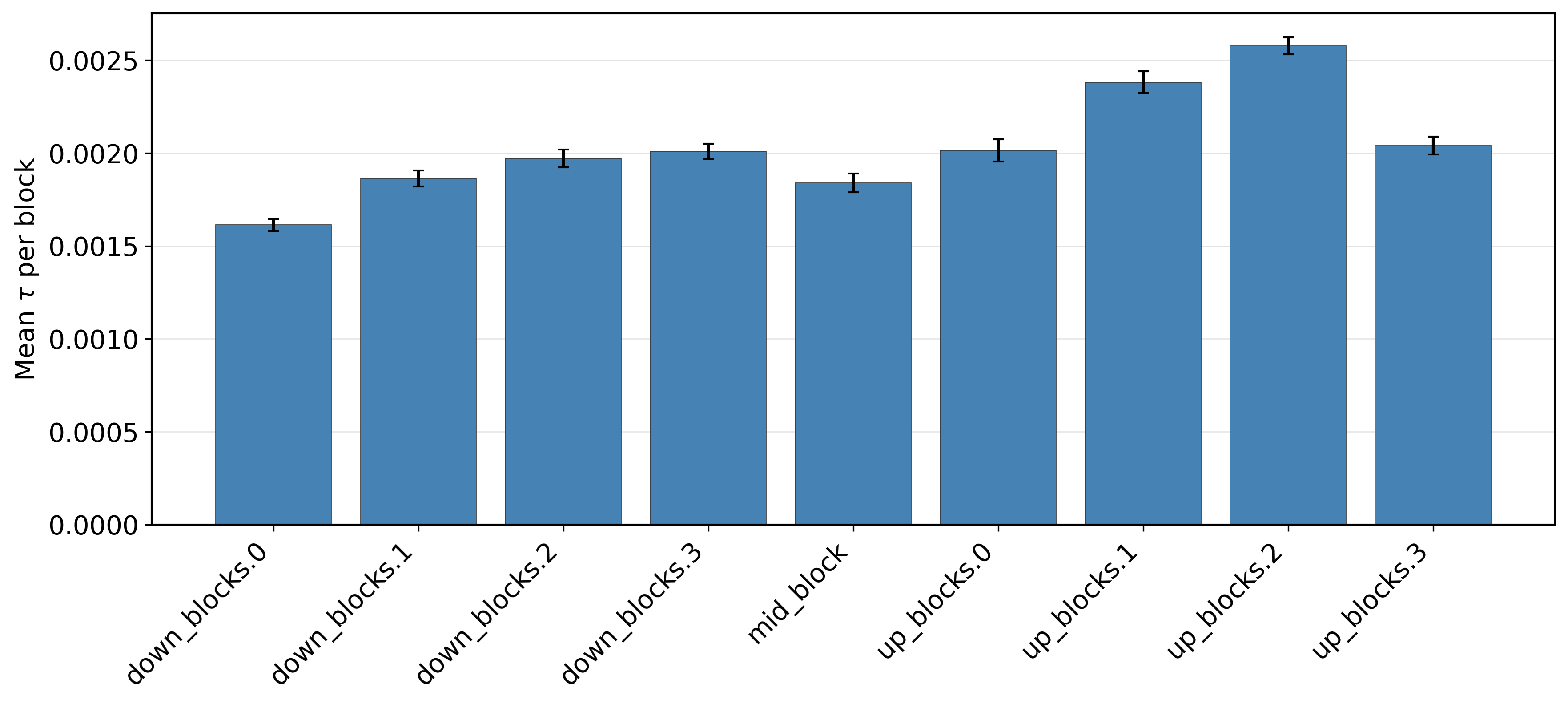}

\caption{\textbf{Per-block shift magnitude.} Cultural vectors produce the largest parameter shifts in \texttt{up-block-1} and \texttt{up-block-2}.}
\label{fig:hist-per-block-shift}

\end{minipage}

\end{figure}














\textbf{Block-wise application of cultural vectors.}
Fig.~\ref{fig:hist-per-block-shift} shows that the largest parameter shifts occur in up-block-1 and up-block-2. Tab.~\ref{tab:tv_layerwise} shows that up-block-1, a relatively deep decoder block, achieves the highest cultural alignment (ALIGN 4.39) and lowest hallucination (HAL 1.28), while maintaining low exaggeration (EXAG 1.88). In contrast, applying cultural vectors to shallower decoder blocks yields lower alignment and higher exaggeration, suggesting that the effectiveness of cultural adaptation depends on the U-Net block.




\subsection{Compositionality of Cultural Vectors}
\label{sec:result_cross_cultural_generation}

\begin{figure}[t]
\centering
\begin{subfigure}[t]{0.49\textwidth}
    \centering
    \includegraphics[width=\linewidth]{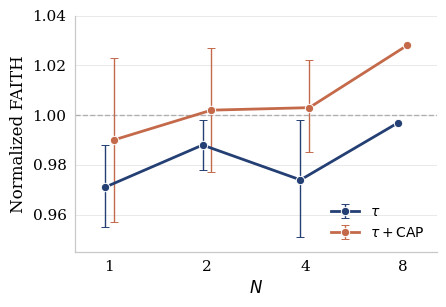}
    \caption{FAITH across all countries.}
    \label{fig:tv_outdomain_b}
\end{subfigure}
\hfill
\begin{subfigure}[t]{0.49\textwidth}
    \centering
    \includegraphics[width=\linewidth]{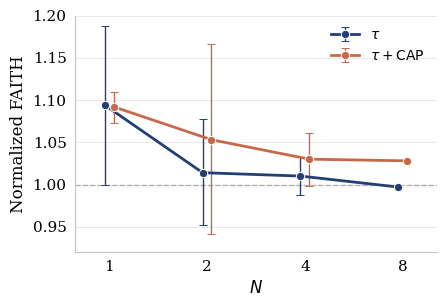}
    \caption{FAITH on seen countries.}
    \label{fig:tv_indomain_c}
\end{subfigure}

\caption{\textbf{Multi-cultural behavior of cultural vectors under composition.}
We uniformly average $N$ randomly selected cultural vectors and evaluate
normalized FAITH on held-out activities.
(a) FAITH across all countries generally increases as more vectors are merged.
(b) FAITH on merged countries decreases with increasing $N$ due to interference.
CAP improves FAITH and compositionality in both settings.
Results are averaged over three random subsets per $N$,
except $N=8$, which has one subset.}
\label{fig:tv_multitask}
\vspace{-6pt}
\end{figure}

Having established that cultural vectors capture meaningful cultural knowledge, we investigate their compositionality in three settings: (1) multi-cultural generation through merging vectors from multiple countries (Fig.~\ref{fig:tv_multitask}); (2) generalization to unseen countries (Tab.~\ref{tab:leave_one_out}); and (3) combining synthetic and real data to improve generation when no real target data is available (Tab.~\ref{tab:gn_gs} in Appendix).

 \begin{table*}[t]
\setlength{\tabcolsep}{3pt}
\centering
\scriptsize
\begin{minipage}[t]{0.50\textwidth}
\caption{\textbf{DGM improves generalization over naive merging.}
DGM improves both TA/TIES. $\alpha=0.7$; \textbf{Bold} denotes best within each method; $\uparrow$/$\downarrow$: higher/lower is better.}
\label{tab:leave_one_out}
\centering

\begin{tabular}{lcccc}
\toprule
\textbf{Method}
& ALIGN $\uparrow$ & EXAG $\downarrow$
& HAL $\downarrow$ & FAITH $\uparrow$ \\
\hline\hline

Pre-trained
& 3.44 & 2.28 & 1.92 & 3.75 \\

\hline

TA~\cite{ilharco2023editing}
& 3.37 & 2.46 & 2.23 & 3.57 \\

TA + DGM
& \textbf{3.51} & \textbf{2.32}
& \textbf{2.03} & \textbf{3.72} \\

\hline

TIES~\cite{yadav2023tiesmerging}
& 3.35 & 2.62 & 2.42 & 3.43 \\

TIES + DGM
& \textbf{3.61} & \textbf{2.19}
& \textbf{1.97} & \textbf{3.82} \\

\bottomrule
\end{tabular}
\end{minipage}
\hfill
\begin{minipage}[t]{0.45\textwidth}

\caption{\textbf{DGM Human evaluation.}
Mean ranks from native annotators across both Global North (GN), Global South (GS) regions. Lower is better. TIES+DGM achieves the lowest mean ranks.}
\label{tab:human_eval}
\centering
\begin{tabular}{lccc}
\toprule
\textbf{Method} & GN & GS & All \\
\hline\hline
Pre-trained & 2.08 & 2.09 & 2.08 \\
\hline
TIES & 2.26 & 2.16 & 2.20 \\
\textbf{TIES+DGM} & \textbf{1.66} & \textbf{1.75} & \textbf{1.72} \\
\bottomrule
\end{tabular}
\end{minipage}

\end{table*}

\textbf{Naive merging reveals cross-cultural interference.}
In Fig.~\ref{fig:tv_multitask}, we evaluate multi-cultural behavior and the compositionality of cultural vectors using Monte Carlo sampling over eight countries from CULTIVate (held-out set activities). For each $N\in\{1,2,4,8\}$, we randomly sample three subsets of $N$ countries (one for $N=8$) and uniformly average their cultural vectors. We evaluate FAITH normalized by each country's individual full fine-tuning baseline, considering both the full country pool and the countries included in each merge. We report mean and stdev across sampled subsets.

\revision{Fig.~\ref{fig:tv_multitask} shows that increasing $N$ generally improves normalized FAITH across the full country pool (Fig.~\ref{fig:tv_outdomain_b}), while reducing faithfulness on the merged countries (Fig.~\ref{fig:tv_indomain_c}), suggesting interference between cultural adaptations. Notably, CAP achieves better merging results than naive cultural vector composition, with performance improvements that persist as the number of merged countries increases, underscoring its effectiveness to improve robustness to cross-cultural interference.}

\textbf{Leave-one-out generalization.}
We evaluate whether cultural vectors from 15 source countries can be composed to generalize to a held-out target country on CULTIVate. To investigate our hypothesis that \textbf{culturally aware composition improves cross-cultural transfer}, we apply DGM to Task Arithmetic (TA)~\cite{ilharco2023editing} and TIES~\cite{yadav2023tiesmerging}.

Tab.~\ref{tab:leave_one_out} shows that naive merging degrades all four metrics relative to the pretrained model. In contrast, \textbf{DGM improves both merging baselines across all metrics}. TIES+DGM achieves the highest FAITH among merging methods (3.82 vs.\ 3.43 for TIES), exceeding the pretrained baseline (3.75), although its HAL remains slightly higher. Human evaluation (Tab.~\ref{tab:human_eval}) further supports these gains: \textbf{TIES+DGM achieves the lowest mean rank} (1.72 vs.\ 2.08 for pretrained and 2.20 for TIES) and ranks first in 10 of 11 countries. These results suggest that descriptor-based cultural similarities provide useful guidance for effective merging.

\begin{figure}[t]
    \centering
\includegraphics[width=0.95\columnwidth]{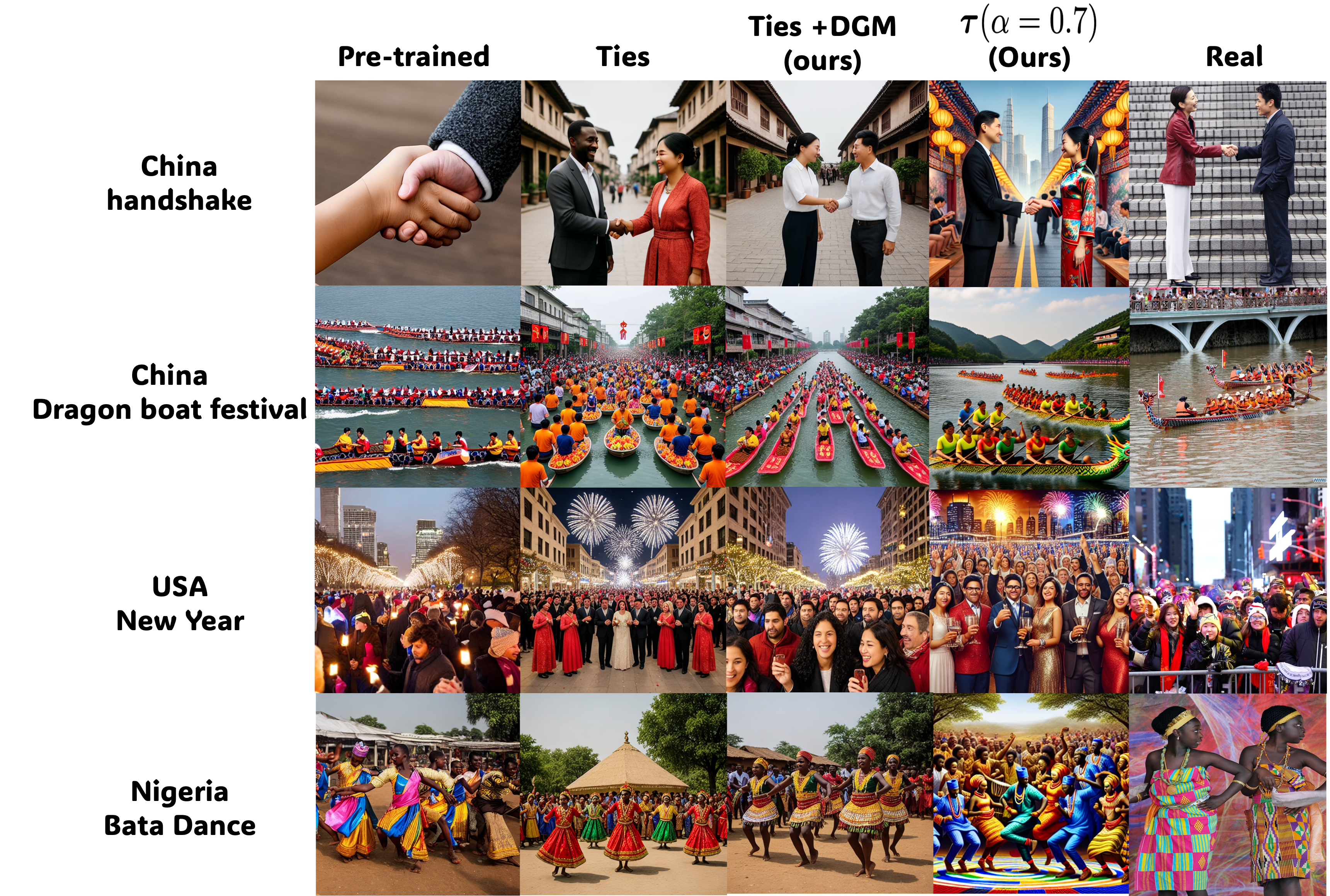}
    \caption{\textbf{Qualitative Examples}. We compare Ties+DGM and $\tv$ against pre-trained and TIES.}
    \label{fig:qualitative_example}
\end{figure}

\textbf{Qualitative example.} 
Fig.~\ref{fig:qualitative_example} compares $\tau$ and TIES+DGM against the pretrained model and TIES~\cite{yadav2023ties} merging. For ``handshake in China,'' both $\tau$ and TIES+DGM produce more culturally grounded depictions than the baselines. For ``dragon boat festival,'' TIES introduces irrelevant flowers, whereas $\tau$ and TIES+DGM better capture the boats associated with the event. Similarly, for ``New Year's in the USA,'' TIES generates an unrelated wedding scene, while $\tau$ and TIES+DGM produce images more consistent with the prompt. These examples illustrate improvements in cultural faithfulness for the selected prompts, complementing the quantitative results.

\section{Conclusion}
\label{sec:conclusion}
In this work, we investigated how cultural knowledge in diffusion models can be represented, controlled, and composed through cultural vectors. We showed that cultural vectors enable controllable culturally grounded generation, with scaling revealing trade-offs between cultural alignment, exaggeration, and hallucination. Our analysis of data sources and internal representations further highlighted the potential of synthetic data for learning cultural representations and shared structure in the activation changes induced by cultural vectors. However, we found that naively merging cultural vectors can lead to cross-cultural interference, limiting their compositionality. To address these challenges, we introduced Cultural Activation Preservation (CAP), a regularization objective that preserves pretrained representations of non-target cultures, and Descriptor-Guided Merging (DGM), which leverages cultural similarities to improve generalization to unseen countries. Overall, our findings provide insights into how cultural knowledge is represented in diffusion models and offer practical approaches toward more controllable and culturally faithful cross-cultural generation.

\section{Limitations}
\label{sec:limitations}

\revision{
First, following prior work~\cite{nayak_culturalframes_2025,bayramli2025diffusion,rege_cure_2025,malakouti2026culture}, we use countries as proxies for cultures, which does not capture within-country diversity. Modeling such diversity requires data that sufficiently captures within cultural differences and expert knowledge, both of which can be difficult to obtain. Although our framework could be extended to finer-grained cultural representations through additional fine-tuning, we leave this direction for future work.
Second, in this work we develop DGM as the simple yet first method to motivate the importance culture-aware compositional methods. DGM relies on descriptor-based cultural similarities; ~\cite{malakouti2026culture} has shown these descriptors can effectively capture cultural activities, but they are still prone to overlook important cultural distinction or reflect LLM biases. Nevertheless, descriptors are inexpensive to generate and require no target-country images, offering a scalable approach to extending cultural adaptation to additional countries and cultural groups.
Finally, although descriptor-guided merging improves cross-cultural generalization, interference between cultural vectors persists. Developing more effective composition methods that preserve individual cultural representations remains an open challenge.
}

\clearpage

\section{Acknowledgment}
\revision{This work is supported by NSF Grant No. 2329992. We gratefully acknowledge the support of those who contributed to the human evaluation. We also thank Aysan Aghazadeh and Yufei Fiona Wang for their valuable comments and help throughout the project.}

\medskip

{
    \small    
    \bibliographystyle{plainnat}
    \bibliography{main}
}

\clearpage
\appendix



\section{Table of Contents}
\begin{itemize}
    \item \textbf{Additional Method Details}
    \begin{itemize}
        \item Cultural descriptor generation
        \item Cross-Cultural Vector Analogy (CVA)
    \end{itemize}

    \item \textbf{Additional Results}
    \begin{itemize}
        \item \textbf{Learning and Controlling Cultural Vectors}
        \begin{itemize}
            \item Cultural faithfulness across countries
            \item Cultural faithfulness across activities
        \end{itemize}

        \item \textbf{Sources and Properties of Cultural Vectors}
        \begin{itemize}
            \item Effect of cultural vector scaling
            \item Parameter-space SVD analysis
            \item Cultural similarity analysis
            \item Activation-space CKA analysis
        \end{itemize}

        \item \textbf{Ablations on Learning Cultural Vectors}
        \begin{itemize}
            \item Impact of synthetic training data diversity
            \item Conditional vs.\ unconditional training
        \end{itemize}

        \item \textbf{Cross-Cultural Generalization}
        \begin{itemize}
            \item Progressive cultural vector merging
            \item CAP and output-preservation ablations
            \item CAP activation-space analysis
            \item Generalization to unseen countries with DGM
            \item Cross-cultural vector analogy
        \end{itemize}

        \item \textbf{Human Evaluation}
        \begin{itemize}
            \item DGM versus pretrained and TIES merging
            \item Real versus synthetic cultural vectors
        \end{itemize}
    \end{itemize}

    \item \textbf{Broader Impact}

    \item \textbf{Evaluation Prompts}
    \begin{itemize}
        \item Cultural alignment (ALIGN)
        \item Cultural exaggeration (EXAG)
        \item Cultural hallucination (HAL)
    \end{itemize}
\end{itemize}

\section{Additional Method Details}
\label{app:method_details}

\textbf{Descriptor Generation.}
Prior work~\cite{Pratt_2023_ICCV,menon2023visual} has shown
that LLM-generated visual descriptors capture object appearances
for zero-shot classification and characterize visual differences
across geographic regions~\cite{Buettner_2024_CVPR,basu2025geodiv}.
Following~\cite{malakouti2026culture}, we use LLM-generated
cultural descriptors across five dimensions: attire, background,
objects, interactions, and spatial layout. Descriptors are
generated by multiple LLMs and consolidated by a separate LLM
to filter inconsistent or incorrect entries. The resulting
descriptor sets $\mathcal{T}_c$ are used to compute cultural
similarities for Descriptor-Guided Merging (DGM), as described
in Sec.~\ref{sec:cc_gen}.

\textbf{Cross-Cultural Vector Analogy (CVA).}
While synthetic data enables scalable learning of cultural
vectors, it may introduce biases inherited from the underlying
text-to-image models. To investigate whether differences between
real and synthetic cultural vectors can be transferred across
countries, we model this difference as an offset in parameter space.

Let $\mathcal{S}\subset\mathcal{C}$ denote the source countries
with both real and synthetic training data. We compute the average
real-to-synthetic offset across source countries and apply it
to the synthetic vector of a target country $t$:
\begin{equation}
\hat{\tv}_{\mathrm{real}}^t =
\tv_{\mathrm{synth}}^t +
\frac{1}{|\mathcal{S}|}
\sum_{s\in\mathcal{S}}
\left(
\tv_{\mathrm{real}}^s -
\tv_{\mathrm{synth}}^s
\right).
\end{equation}
We report the results in Tab.~\ref{tab:gn_gs}.


\section{Additional Results}
\label{app:additional_results}

\begin{table*}[t]
\setlength{\tabcolsep}{1pt}
\renewcommand{\arraystretch}{1.05}
\caption{\textbf{Controlling cultural adaptation via cultural vectors on CULTIVate.} 
We vary the scaling coefficient $\alpha$, and all vectors are applied to the full model. 
$\uparrow$/$\downarrow$ indicate higher/lower is better. 
Results are mean \pmsd\ standard deviation (GPT-4o-as-a-judge). Region's average is sample weighted.}
\centering
\scriptsize
\begin{tabular}{l||cccc|cccc|cccc}
\toprule
\multirow{2}{*}{\textbf{Method}}
& \multicolumn{4}{c|}{\textbf{Global North}} 
& \multicolumn{4}{c|}{\textbf{Global South}} 
& \multicolumn{4}{c}{\textbf{All}} \\
& ALIGN $\uparrow$ & EXAG $\downarrow$ & HAL $\downarrow$ & FAITH $\uparrow$
& ALIGN $\uparrow$ & EXAG $\downarrow$ & HAL $\downarrow$ & FAITH $\uparrow$
& ALIGN $\uparrow$ & EXAG $\downarrow$ & HAL $\downarrow$ & FAITH $\uparrow$ \\
\hline
\hline

Pre-trained 

& 3.67 \std{0.21} & 1.98 \std{0.26} & 1.64 \std{0.26} & 4.02
& 3.33 \std{0.33} & 2.42 \std{0.23} & 2.04 \std{0.39} & 3.62
& 3.44 \std{0.33} & 2.28 \std{0.31} & 1.92 \std{0.39} & 3.75 \\
\hline

$\tau$ ($\alpha=0.3$)
& 4.17 \std{0.29} & 1.74 \std{0.17} & 1.50 \std{0.19} & 4.31
& 3.99 \std{0.35} & 1.95 \std{0.24} & 1.59 \std{0.34} & 4.15
& 4.05 \std{0.33} & 1.88 \std{0.23} & 1.56 \std{0.30} & 4.20 \\

$\tau$ ($\alpha=0.5$)
& 4.39 \std{0.18} & \textbf{1.73} \std{0.18} & 1.29 \std{0.14} & 4.46
& 4.29 \std{0.34} & \textbf{1.89} \std{0.21} & 1.31 \std{0.24} & 4.36
& 4.32 \std{0.30} & \textbf{1.84} \std{0.21} & 1.31 \std{0.21} & 4.39 \\

$\tau$ ($\alpha=0.7$)
& \textbf{4.55} \std{0.16} & 1.83 \std{0.19} & \textbf{1.20} \std{0.10} & \textbf{4.51}
& \textbf{4.56} \std{0.25} & 2.07 \std{0.25} & \textbf{1.17} \std{0.10} & \textbf{4.44}
& \textbf{4.56} \std{0.22} & 2.00 \std{0.26} & \textbf{1.18} \std{0.10} & \textbf{4.46} \\

$\tau$ ($\alpha=2.0$)
& 1.28 \std{0.19} & 4.66 \std{0.13} & 4.09 \std{0.48} & 1.51
& 1.43 \std{0.18} & 4.58 \std{0.17} & 3.96 \std{0.25} & 1.63
& 1.38 \std{0.19} & 4.61 \std{0.16} & 4.00 \std{0.33} & 1.59 \\
\hline

Fine-tuned 
& 4.50 \std{0.10} & 2.30 \std{0.15} & 1.29 \std{0.11} & 4.30
& 4.48 \std{0.30} & 2.43 \std{0.26} & 1.25 \std{0.19} & 4.27
& 4.49 \std{0.25} & 2.39 \std{0.24} & 1.26 \std{0.17} & 4.28 \\

\bottomrule
\end{tabular}
\label{tab:full_tv_alpha}
\end{table*}

\begin{figure*}[t]
\centering

\begin{minipage}[c]{0.47\textwidth}
    \centering
\captionof{table}{
\textbf{Effect of cultural vector scaling on CulturalFrame.}
We report ALIGN, EXAG, HAL, and FAITH across different scaling coefficients $\alpha$. Results are mean \pmsd\ standard deviation (GPT-4o-as-a-judge).
}
    \label{tab:CF_tv_alpha_ablation}
    \vspace{4pt}
    \scriptsize
    \setlength{\tabcolsep}{1.5pt}

    \begin{tabular}{lcccc}
    \toprule
    \textbf{Method}
    & ALIGN $\uparrow$
    & EXAG $\downarrow$
    & HAL $\downarrow$
    & FAITH $\uparrow$ \\
    \midrule
    Pre-trained
    & 3.91 \std{0.15} & 1.75 \std{0.21}
    & 1.38 \std{0.15} & 4.26 \\
    \midrule
    $\tv$ ($\alpha=0.3$)
    & 4.05 \std{0.11} & \textbf{1.71} \std{0.22}
    & 1.33 \std{0.13} & 4.34 \\

    $\tv$ ($\alpha=0.5$)
    & 4.15 \std{0.14} & 1.73 \std{0.22}
    & 1.29 \std{0.11} & 4.38 \\

    $\tv$ ($\alpha=0.7$)
    & 4.16 \std{0.11} & 1.73 \std{0.24}
    & \textbf{1.25} \std{0.10} & \textbf{4.39} \\

    $\tv$ ($\alpha=2.0$)
    & 4.01 \std{0.13} & 2.54 \std{0.25}
    & 1.36 \std{0.17} & 4.04 \\
    \midrule
    Fine-tuned
    & \textbf{4.19} \std{0.08} & 1.85 \std{0.25}
    & 1.28 \std{0.10} & 4.35 \\
    \bottomrule
    \end{tabular}
\end{minipage}
\hspace{0.01\textwidth}
\begin{minipage}[c]{0.50\textwidth}
    \centering

    \captionof{figure}{
        \textbf{Scaling trade-offs.}
        Cultural alignment versus exaggeration and hallucination
        under different scaling coefficients $\alpha$.
    }
    \label{fig:tv_scaling_tradeoffs}

    \begin{subfigure}[t]{0.49\linewidth}
        \centering
        \includegraphics[width=\linewidth]{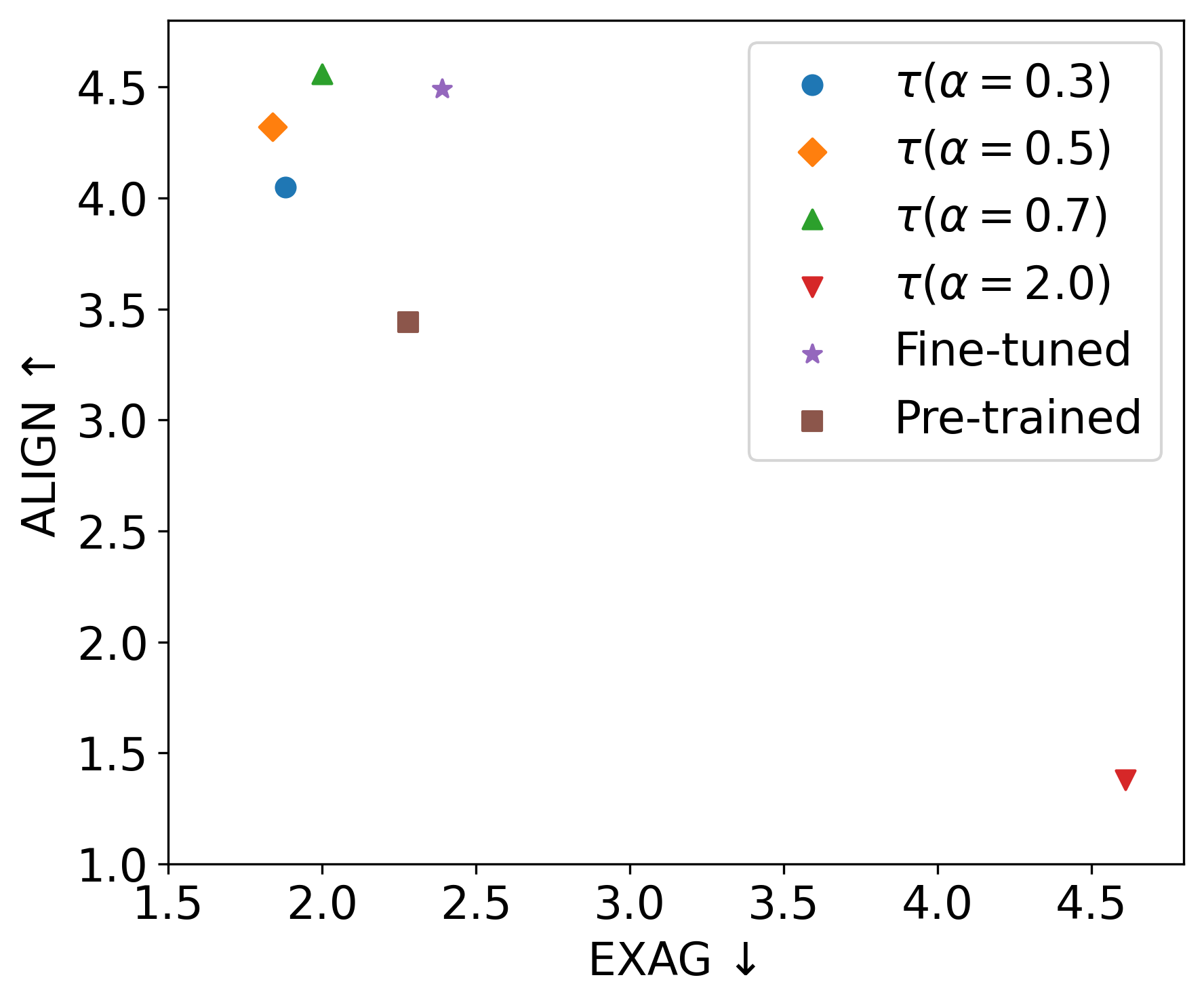}
        \caption{ALIGN vs.\ EXAG}
        \label{fig:ALIGN_vs_EXAG}
    \end{subfigure}
    \hfill
    \begin{subfigure}[t]{0.49\linewidth}
        \centering
        \includegraphics[width=\linewidth]{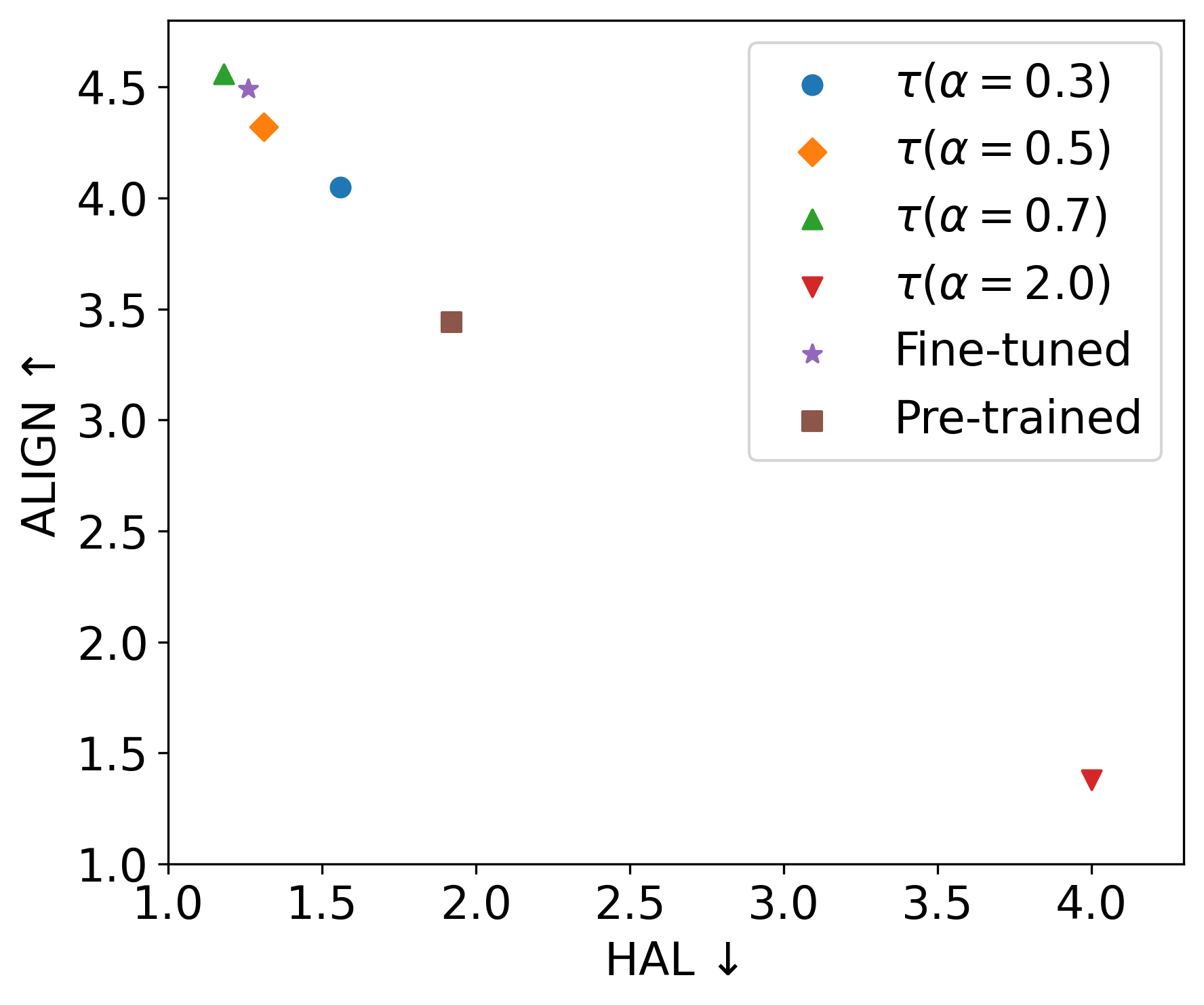}
        \caption{ALIGN vs.\ HAL}
        \label{fig:ALIGN_vs_HAL}
    \end{subfigure}
\end{minipage}
\vspace{-5pt}
\end{figure*}










\begin{table*}[t]
\centering
\caption{\textbf{Cultural faithfulness across Global South (GS)
and Global North (GN) countries.}
FAITH scores for the pretrained model, full fine-tuning
(Full-FT), and cultural vectors ($\alpha=0.7$).
Relative changes over Full-FT are shown in parentheses.
Green and red indicate improvements and declines,
respectively. Best scores are in \textbf{bold}.}
\label{tab:country_faithfulness}

\scriptsize
\setlength{\tabcolsep}{3pt}
\renewcommand{\arraystretch}{1.08}

\begin{minipage}[t]{0.53\textwidth}
\centering
\textbf{Global South (GS)}

\vspace{3pt}
\begin{tabular}{lccc}
\toprule
\textbf{Country} & \textbf{Pretrained}
& \textbf{Full-FT} & $\boldsymbol{\tv}$ \\
\midrule
Brazil & 3.60 & \textbf{4.56}
& \textbf{4.56} {\tiny\color{gray}(0.0\%)} \\
China & 3.97 & 4.27
& \textbf{4.43} {\tiny\color{green!50!black}(+3.7\%)} \\
India & 4.11 & \textbf{4.44}
& \textbf{4.44} {\tiny\color{gray}(0.0\%)} \\
Indonesia & 3.48 & 4.22
& \textbf{4.56} {\tiny\color{green!50!black}(+8.1\%)} \\
Iran & 3.15 & 3.81
& \textbf{4.11} {\tiny\color{green!50!black}(+7.9\%)} \\
Mexico & 3.87 & 4.31
& \textbf{4.62} {\tiny\color{green!50!black}(+7.2\%)} \\
Nepal & 3.89 & 4.29
& \textbf{4.38} {\tiny\color{green!50!black}(+2.1\%)} \\
Nigeria & 3.74 & \textbf{4.31}
& 4.27 {\tiny\color{red!70!black}(-0.9\%)} \\
Philippines & 3.45 & 4.56
& \textbf{4.68} {\tiny\color{green!50!black}(+2.6\%)} \\
South Africa & 3.59 & 4.18
& \textbf{4.37} {\tiny\color{green!50!black}(+4.5\%)} \\
Turkey & 3.30 & 4.01
& \textbf{4.40} {\tiny\color{green!50!black}(+9.7\%)} \\
\midrule
\textbf{GS Average} & 3.65 & 4.27
& \textbf{4.44} {\tiny\color{green!50!black}(+4.0\%)} \\
\bottomrule
\end{tabular}
\end{minipage}
\hfill
\begin{minipage}[t]{0.43\textwidth}
\centering
\textbf{Global North (GN)}

\vspace{3pt}
\begin{tabular}{lccc}
\toprule
\textbf{Country} & \textbf{Pretrained}
& \textbf{Full-FT} & $\boldsymbol{\tv}$ \\
\midrule
France & 4.17 & 4.38
& \textbf{4.40} {\tiny\color{green!50!black}(+0.5\%)} \\
Germany & 3.94 & 4.25
& \textbf{4.57} {\tiny\color{green!50!black}(+7.5\%)} \\
Italy & 4.12 & 4.27
& \textbf{4.54} {\tiny\color{green!50!black}(+6.3\%)} \\
Spain & 3.64 & 4.22
& \textbf{4.48} {\tiny\color{green!50!black}(+6.2\%)} \\
USA & 4.23 & 4.42
& \textbf{4.56} {\tiny\color{green!50!black}(+3.2\%)} \\
\midrule
\textbf{GN Average} & 4.02 & 4.31
& \textbf{4.51} {\tiny\color{green!50!black}(+4.6\%)} \\
\bottomrule
\end{tabular}
\end{minipage}
\end{table*}

\begin{table*}[t]
\centering
\caption{\textbf{Cultural faithfulness across activity categories.} We report FAITH scores for the pretrained model, full fine-tuning (Full-FT), and cultural vectors ($\alpha=0.7$). Relative improvements over Full-FT are shown in parentheses, with \textcolor{green!50!black}{green} indicating improvements and \textcolor{red!70!black}{red} indicating declines. Best scores are in bold.}
\label{tab:activity_faithfulness}
\scriptsize
\setlength{\tabcolsep}{3pt}
\begin{tabular}{lccccccccc}
\toprule
\textbf{Model}
& \textbf{Celebration}
& \textbf{Concert}
& \textbf{Dancing}
& \textbf{Eating}
& \textbf{Funeral}
& \textbf{Game}
& \textbf{Greeting}
& \textbf{Religious}
& \textbf{Wedding} \\
\hline
\hline
Pretrained
& 3.82 & 4.39 & 3.74 & 3.91 & 4.24
& 2.89 & 4.19 & 4.03 & 4.19 \\
Full-FT
& 4.28 & 4.27 & 4.44 & 4.12 & 4.07
& 4.22 & 4.30 & 4.27 & 4.26 \\
\hline
\hline
$\tv$ ($\alpha=0.7$)
& \textbf{4.47} {\tiny\color{green!50!black}(+4.4\%)}
& \textbf{4.44} {\tiny\color{green!50!black}(+4.0\%)}
& \textbf{4.50} {\tiny\color{green!50!black}(+1.4\%)}
& \textbf{4.47} {\tiny\color{green!50!black}(+8.5\%)}
& \textbf{4.25} {\tiny\color{green!50!black}(+4.4\%)}
& \textbf{4.44} {\tiny\color{green!50!black}(+5.2\%)}
& \textbf{4.49} {\tiny\color{green!50!black}(+4.4\%)}
& \textbf{4.43} {\tiny\color{green!50!black}(+3.7\%)}
& \textbf{4.43} {\tiny\color{green!50!black}(+4.0\%)} \\
\bottomrule
\end{tabular}
\vspace{-15pt}
\end{table*}

\subsection{Learning and Controlling Cultural Vectors}
\label{app:learning_vectors}

\textbf{Performance across Global South and Global North countries.}
As shown in Tab.~\ref{tab:full_tv_alpha}, cultural vectors
with $\alpha=0.7$ improve cultural alignment (ALIGN) over
full fine-tuning across both Global South (GS) and Global
North (GN) countries, while reducing cultural exaggeration
(EXAG). These results suggest that scaling cultural vectors
provides a more favorable balance between alignment and
exaggeration than full fine-tuning.

\revision{
We further examine country-level cultural faithfulness (FAITH)
in Tab.~\ref{tab:country_faithfulness}. Cultural vectors
outperform or match full fine-tuning in 15 of 16 countries,
with average relative improvements of 4.0\% and 4.6\%
across GS and GN countries, respectively.
}

\revision{
\textbf{Performance across cultural activities.}
To examine whether the improvements extend across different
cultural scenarios, we compare cultural faithfulness across
nine activity categories. As shown in
Tab.~\ref{tab:activity_faithfulness}, cultural vectors
outperform full fine-tuning in all nine categories,
with the largest relative improvement for eating (+8.5\%).
These results indicate that the improvements are not
restricted to particular countries or activity categories.
}



\begin{figure}[t]
    \centering
    \setlength{\fboxsep}{0pt}
    \begin{subfigure}[b]{0.32\textwidth}
        \includegraphics[width=\linewidth]{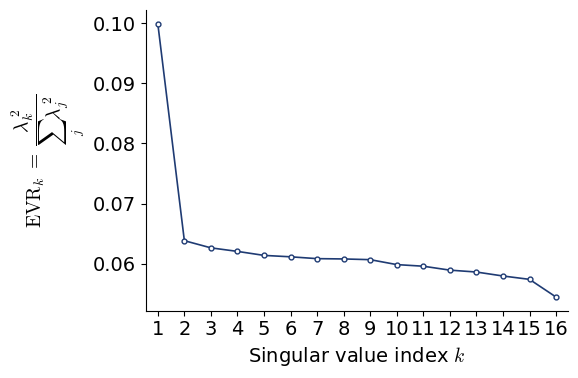}
        \caption{Independence of cultural vectors}
        \label{fig:svd_vecto_appendix}
    \end{subfigure}
    \hfill
    \begin{subfigure}[b]{0.32\textwidth}
        \includegraphics[width=\linewidth]{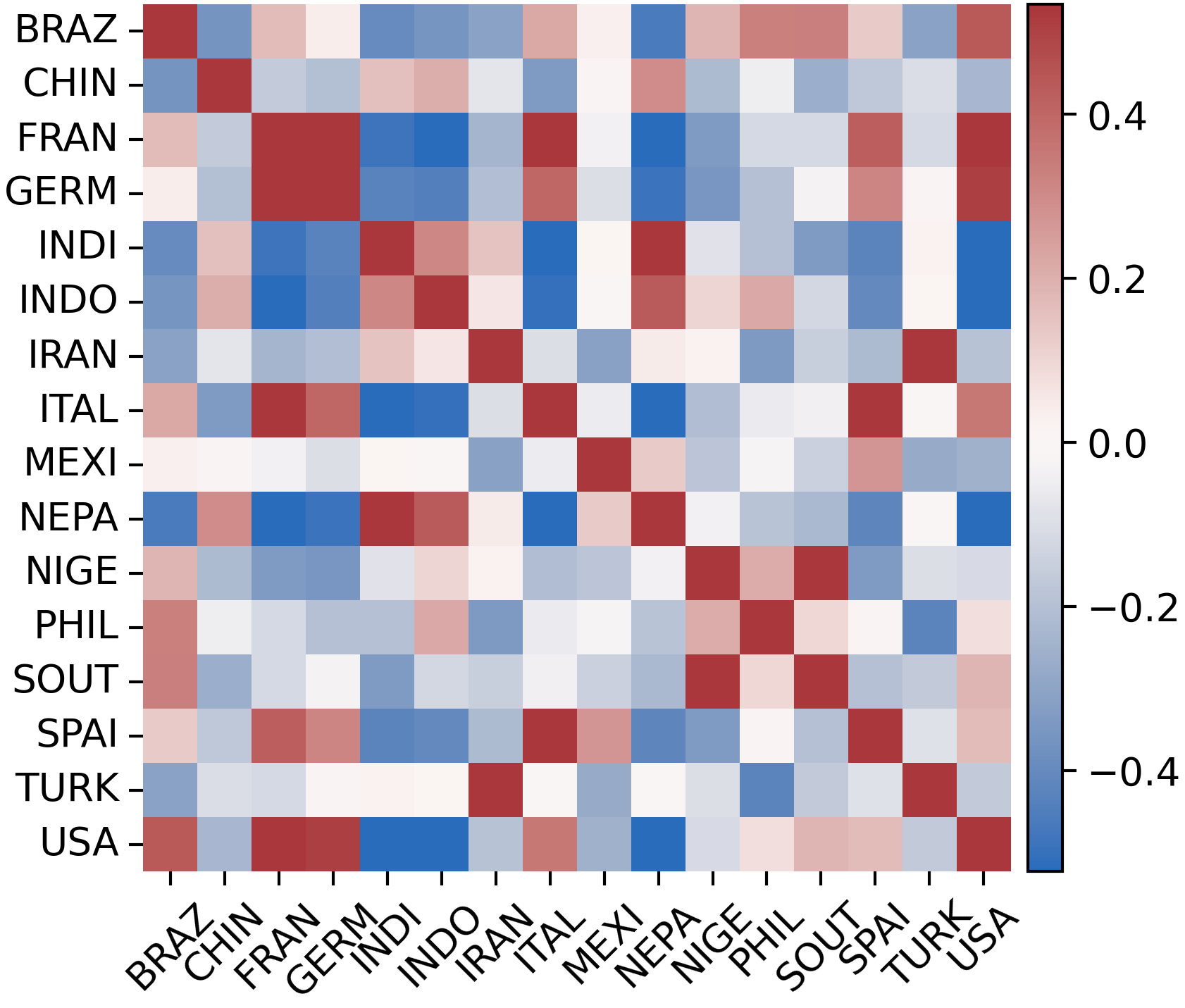}
        \caption{Descriptor similarity}
        \label{fig:heatmap_descriptors}
    \end{subfigure}
    \hfill
    \begin{subfigure}[b]{0.32\textwidth}
        \includegraphics[width=\linewidth]{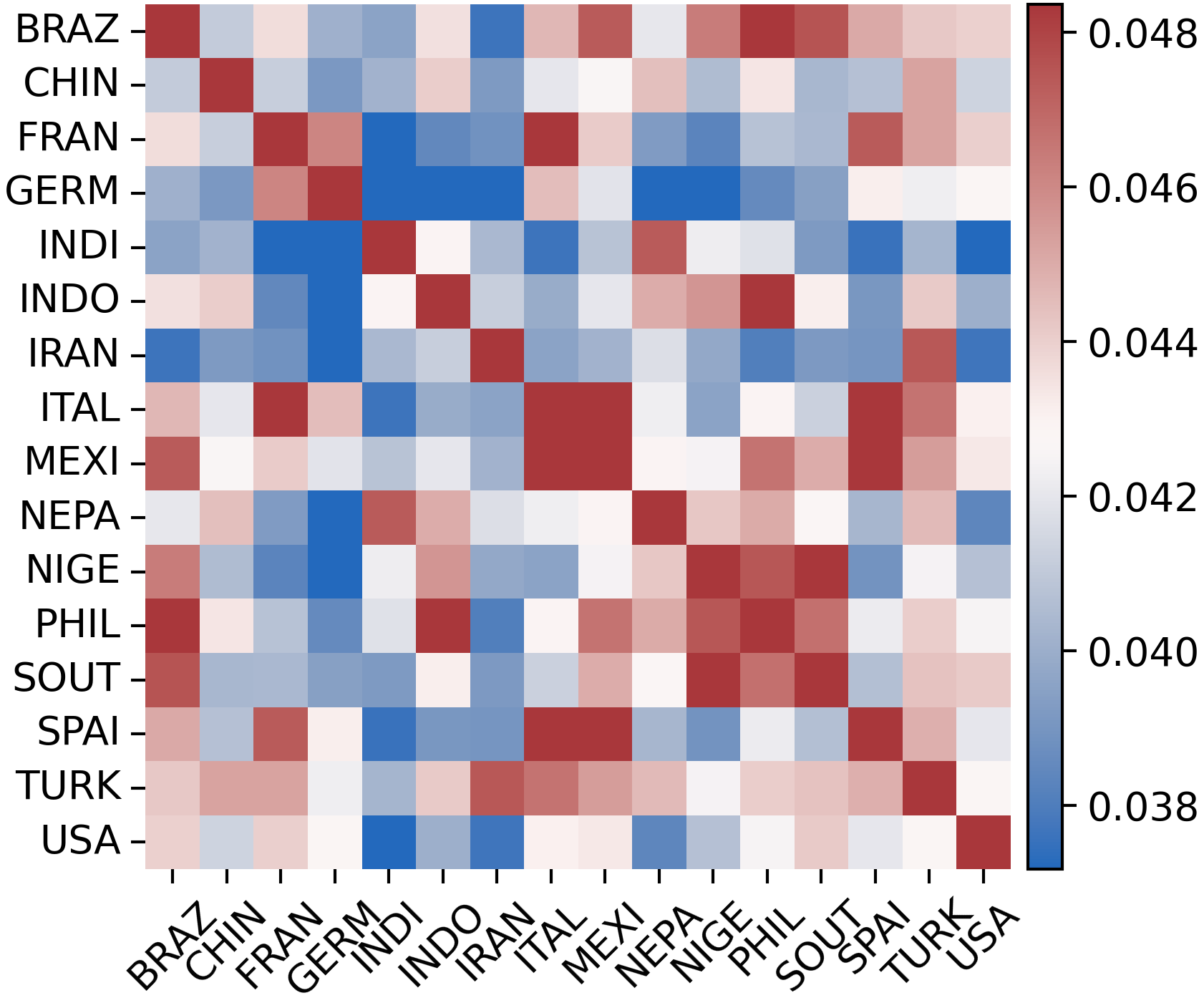}
        \caption{Cultural vector similarity}
        \label{fig:heatmap_tv}
    \end{subfigure}
    \caption{\textbf{Cultural vectors reflect cultural similarity.} (a) Cultural vectors occupy largely independent directions in parameter space, as shown by the distributed singular value spectrum. (b,c) Descriptor and cultural vector similarity matrices, scaled to the 5-95th percentiles to enhance contrast.}
\end{figure}

\begin{figure}[t]
    \centering

    \begin{subfigure}[b]{0.32\textwidth}
        \centering
        \includegraphics[width=\linewidth]{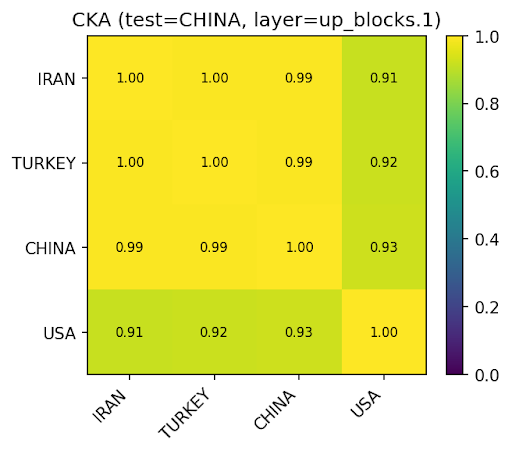}
        \caption{CKA China}
        \label{fig:cka_cn}
    \end{subfigure}
    \hfill
    \begin{subfigure}[b]{0.32\textwidth}
        \centering
        \includegraphics[width=\linewidth]{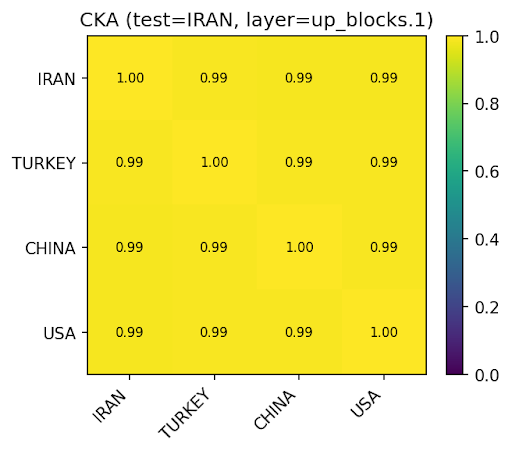}
        \caption{CKA on Iran}
        \label{fig:cka_ir}
    \end{subfigure}
    \hfill
    \begin{subfigure}[b]{0.32\textwidth}
        \centering
        \includegraphics[width=\linewidth]{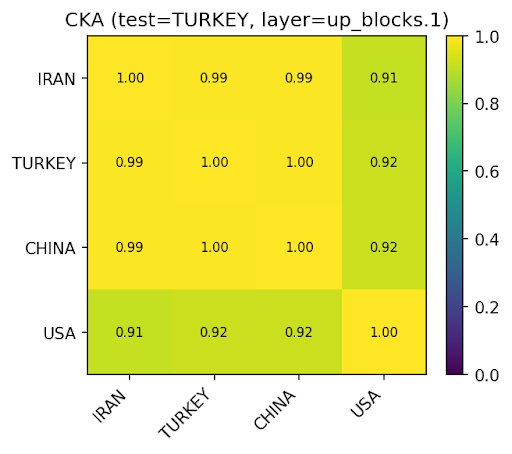}
        \caption{CKA on Turkey}
        \label{fig:cka_tr}
    \end{subfigure}

    \caption{CKA~\cite{kornblith2019similarity} analysis.}
    \label{fig:three_subfigures}
\vspace{-15pt}
\end{figure}

\subsection{Sources and Properties of Cultural Vectors}
\label{app:properties_vectors}

\textbf{Impact of scaling $\alpha$.}
Tabs.~\ref{tab:full_tv_alpha} and
\ref{tab:CF_tv_alpha_ablation} examine the effect of
varying the scaling coefficient. Smaller values of $\alpha$
generally yield lower cultural alignment and exaggeration,
while larger values strengthen cultural adaptation.
However, excessive scaling (e.g., $\alpha=2$) can degrade
image quality.

Figs.~\ref{fig:ALIGN_vs_EXAG} and
\ref{fig:ALIGN_vs_HAL} further illustrate that increasing
$\alpha$ tends to improve cultural alignment while also
increasing exaggeration and hallucination. Intermediate
scaling values provide a more favorable balance between
these objectives.

\textbf{Cultural vectors occupy independent directions in parameter space}. As shown in Fig.~\ref{fig:svd_vecto_appendix}, the explained variance is distributed nearly uniformly across singular values for up-block-1, yielding effective ranks of 15.83 out of a maximum of 16. This near full-rank structure indicates that cultural vectors span largely independent directions in parameter space, rather than concentrating in a shared low-dimensional subspace.

\revision{
\textbf{Cultural similarity.}
Fig.~\ref{fig:similarity_corr} shows a positive correlation
between pairwise cultural vector similarities and
descriptor-based cultural similarities. The corresponding
similarity matrices (Figs.~\ref{fig:heatmap_descriptors}
and~\ref{fig:heatmap_tv}) exhibit related patterns across
countries. These observations suggest that parameter-space
similarities reflect some of the relationships captured
by cultural descriptors, although descriptor similarity
is only a proxy for cultural relatedness.
}

\textbf{CKA analysis}. Following
~\cite{stoica2025knots}, we apply centered kernel alignment
(CKA)~\cite{kornblith2019similarity} to compare activations
across countries
(Figs.~\ref{fig:cka_cn}, \ref{fig:cka_ir}, and
\ref{fig:cka_tr}). We observe high representation similarity
across the analyzed countries, including pairs with
different cultural descriptors. The high similarities indicate shared activation structure across vectors.



\begin{table*}[t]
\centering

\begin{minipage}[t]{0.49\textwidth}
    \vspace{0pt}
    \centering
    \captionof{table}{
    \textbf{Effect of training data diversity on cultural vectors.}
    Comparison of cultural vectors ($\tv$) and full fine-tuning
    using synthetic data from different T2I models.
    Values denote mean scores.
    }
    \label{tab:ablation_num_T2I}

    \vspace{4pt}
    \scriptsize
    \setlength{\tabcolsep}{1pt}
    
    \begin{tabular}{llcccc}
    \toprule
    \textbf{Training Data} & \textbf{Model} & $\alpha$
    & HAL $\downarrow$ & EXAG $\downarrow$ & ALIGN $\uparrow$ \\
    \midrule
    Pre-trained & -- & -- & 1.92 & 2.28 & 3.44 \\
    \midrule
    \multirow{2}{*}{\shortstack[l]{3 T2I Models\\(Private)}}
    & $\tv$ & 0.7 & 1.18 & 2.00 & 4.56 \\
    & Fine-Tuned & -- & 1.26 & 2.39 & 4.49 \\
    \midrule
    \multirow{2}{*}{\shortstack[l]{6 T2I Models\\(Private + Public)}}
    & $\tv$ & 0.7 & 1.40 & 2.18 & 4.26 \\
    & Fine-Tuned & -- & 1.39 & 2.54 & 4.24 \\
    \bottomrule
    \end{tabular}
\end{minipage}
\hfill
\begin{minipage}[t]{0.49\textwidth}
    \vspace{0pt}
    \centering
    \captionof{table}{
    \textbf{Conditional vs.\ unconditional cultural adaptation.}
    Comparison of cultural vectors ($\tv$) and full fine-tuning
    under conditional and unconditional training.
    Values denote mean scores.
    }
    \label{tab:conditional_unconditional}

    \vspace{4pt}
    \scriptsize
    \setlength{\tabcolsep}{1pt}

    \begin{tabular}{llcccc}
    \toprule
    \textbf{Training} & \textbf{Model} & $\alpha$
    & HAL $\downarrow$ & EXAG $\downarrow$ & ALIGN $\uparrow$ \\
    \midrule
    Pre-trained & -- & -- & 1.92 & 2.28 & 3.44 \\
    \midrule
    \multirow{2}{*}{Conditional}
    & $\tv$ & 0.7 & 1.18 & 2.00 & 4.56 \\
    & Fine-Tuned & -- & 1.26 & 2.39 & 4.49 \\
    \midrule
    \multirow{2}{*}{Unconditional}
    & $\tv$ & 0.7 & 1.80 & 2.64 & 3.60 \\
    & Fine-Tuned & -- & 2.39 & 3.46 & 2.99 \\
    \bottomrule
    \end{tabular}
\end{minipage}

\end{table*}














\subsection{Ablations on Learning Cultural Vectors}

\textbf{Impact of the number of T2I models used to learn cultural vectors.}
In Tab.~\ref{tab:ablation_num_T2I}, we investigate the effect
of increasing the number of T2I models used to generate
synthetic training data. Following the experimental setup
in Sec.~\ref{sec:exp_setup}, we additionally include synthetic
data generated by three public T2I models (Stable Diffusion
3 Medium, FLUX.1, and Qwen-Image), increasing the number of
source models from three to six.

We observe that in both settings (i.e., 3 and 6 T2I model), cultural vectors significantly outperform pre-trained model while improve full finetuned baseline. This further confirms the effectiveness of cultural vectors in steering the generation of models towards more faithful cultural images. However, we observe including more models’ and increasing the training data size, performance degrades compared with three proprietary T2I models as training data sources. We hypothesize that this is due to additional noise and more stereotypical or culturally inaccurate images by public models compared to strong proprietary models.

\textbf{Impact of training strategy.}
Tab.~\ref{tab:conditional_unconditional} shows that cultural
vectors outperform the pretrained baseline under both
conditional and unconditional training. Conditional training
yields stronger absolute performance (ALIGN 4.56 vs. 3.60),
while the relative gain of cultural vectors over full
fine-tuning is larger under unconditional training.


\subsection{Cross-Cultural Generalization}
\label{app:cultural_generalization}

\begin{table}[t]
\caption{\textbf{Effect of progressively adding cultural vectors.} 
Values show FAITH. Values in parentheses indicate change relative to the pretrained model (\textcolor{green!50!black}{improvement}, \textcolor{red}{degradation}). 
$Avg$ is average. $\tau_{IR}$, $\tau_{US}$, and $\tau_{CN}$ denote cultural vectors of Iran, United States, and China, respectively.}

\centering
\scriptsize
\label{tab:tv_progressive}

\begin{tabular}{l||cccc}
\toprule
\textbf{Cultural Vectors Added} 
& \textbf{China} 
& \textbf{Iran} 
& \textbf{Turkey} 
& \textbf{USA} \\
\hline
\hline

Pre-trained 
& 3.79
& 3.31
& 2.64
& 3.92 \\
\hline

$+\, \tau_{IR}$ 
& 2.80 { \textcolor{red}{(-0.99)}}
& 3.89 { \textcolor{green!50!black}{(+0.58)}}
& 3.12 { \textcolor{green!50!black}{(+0.48)}}
& 2.48 { \textcolor{red}{(-1.44)}} \\

$+\, Avg(\tau_{IR}, \tau_{US})$
& 3.88 { \textcolor{green!50!black}{(+0.09)}}
& 3.73 { \textcolor{green!50!black}{(+0.42)}}
& 3.16 { \textcolor{green!50!black}{(+0.52)}}
& 4.23 { \textcolor{green!50!black}{(+0.31)}} \\

$+\, Avg(\tau_{IR}, \tau_{US}, \tau_{CN})$
& 3.82 { \textcolor{green!50!black}{(+0.03)}}
& 3.45 { \textcolor{green!50!black}{(+0.14)}}
& 2.78 { \textcolor{green!50!black}{(+0.14)}}
& 4.30 { \textcolor{green!50!black}{(+0.38)}} \\

\bottomrule
\end{tabular}
\vspace{-15pt}
\end{table}

\begin{table}[t]
\centering
\caption{\textbf{Ablation of preservation objectives and regularization strength.}
We compare CAP and output preservation across different $\lambda$ values and cultural vector compositions. ALIGN is averaged over four countries (Iran, USA, China, and Turkey). Multiple vectors are uniformly averaged. Parentheses indicate changes from pretrained (\textcolor{green!50!black}{improvement}, \textcolor{red!70!black}{degradation}). Best scores are in \textbf{bold}.}
\label{tab:cap_ablation}
\scriptsize
\setlength{\tabcolsep}{3pt}
\begin{tabular}{ll||ccc}
\toprule
\textbf{Method} & $\boldsymbol{\lambda}$
& $\boldsymbol{\tv^{IR}}$
& $\boldsymbol{\mathrm{Avg}(\tv^{IR},\tv^{US})}$
& $\boldsymbol{\mathrm{Avg}(\tv^{IR},\tv^{US},\tv^{CN})}$ \\
\hline
\hline
Pretrained & --
& 3.42 & 3.42 & 3.42 \\
$\tv$ (No Preservation) & --
& 2.87 {\tiny\color{red!70!black}(-0.55)}
& 3.50 {\tiny\color{green!50!black}(+0.08)}
& 3.41 {\tiny\color{red!70!black}(-0.01)} \\
\hline
\hline
Output Preservation
& 0.01
& 3.04 {\tiny\color{red!70!black}(-0.38)}
& 3.40 {\tiny\color{red!70!black}(-0.02)}
& 3.52 {\tiny\color{green!50!black}(+0.10)} \\
& 0.1
& 3.12 {\tiny\color{red!70!black}(-0.30)}
& 3.36 {\tiny\color{red!70!black}(-0.06)}
& 3.39 {\tiny\color{red!70!black}(-0.03)} \\
& 1.0
& 3.53 {\tiny\color{green!50!black}(+0.11)}
& 3.66 {\tiny\color{green!50!black}(+0.24)}
& 3.80 {\tiny\color{green!50!black}(+0.38)} \\
\hline
\hline
CAP
& 0.01
& 3.68 {\tiny\color{green!50!black}(+0.26)}
& 3.69 {\tiny\color{green!50!black}(+0.27)}
& 3.65 {\tiny\color{green!50!black}(+0.23)} \\
& 0.1
& \textbf{3.73} {\tiny\color{green!50!black}(+0.31)}
& 3.75 {\tiny\color{green!50!black}(+0.33)}
& \textbf{3.85} {\tiny\color{green!50!black}(+0.43)} \\
& 1.0
& 3.60 {\tiny\color{green!50!black}(+0.18)}
& \textbf{3.89} {\tiny\color{green!50!black}(+0.47)}
& 3.77 {\tiny\color{green!50!black}(+0.35)} \\
\bottomrule
\end{tabular}
\end{table}

\begin{figure*}[t]
\centering

\begin{subfigure}[t]{0.32\textwidth}
    \centering
    \includegraphics[width=\linewidth]{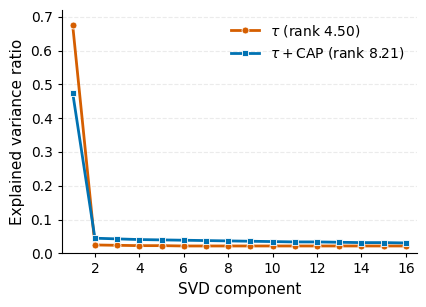}
    \caption{real images}
    \label{fig:evr-cap-real-images}
\end{subfigure}
\hfill
\begin{subfigure}[t]{0.32\textwidth}
    \centering
    \includegraphics[width=\linewidth]{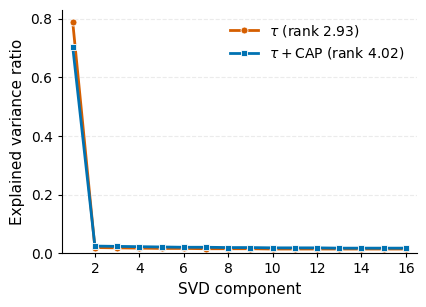}
    \caption{Synthetic data from private models (train data)}
    \label{fig:evr_cap_private_t2i}
\end{subfigure}
\hfill
\begin{subfigure}[t]{0.32\textwidth}
    \centering
    \includegraphics[width=\linewidth]{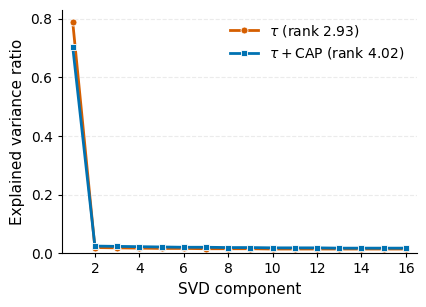}
    \caption{Synthetic data from public T2I}
    \label{fig:evr_cap_public_t2i}
\end{subfigure}
\caption{\textbf{Activation-space Spectrum.} 
SVD analysis (at 400 time-step) show that CAP increases the rank of the activations across different images: real (unseen), private T2I (trained data), and public T2I (unseen). }
\vspace{-15pt}
\end{figure*}

\begin{table*}[t]
\centering
\caption{\textbf{Descriptor-Guided Merging (DGM) improves cross-cultural generalization in the leave-one-out setting.}
Source cultural vectors are merged to generate images for a held-out target culture with $\alpha=0.7$ on CULTIVate (16 countries).
Direct-adaptation methods use target-country training data and are included only as references.
Bold indicates the best result within each method group;
$\uparrow$/$\downarrow$ indicate higher/lower is better.}
\label{tab:leave_one_out_full}

\scriptsize
\setlength{\tabcolsep}{1pt}

\begin{tabular}{ll||cccc|cccc|cccc}
\toprule
\multirow{2}{*}{\textbf{Group}}
& \multirow{2}{*}{\textbf{Method}}
& \multicolumn{4}{c|}{\textbf{Global North}}
& \multicolumn{4}{c|}{\textbf{Global South}}
& \multicolumn{4}{c}{\textbf{All}} \\
&
& ALIGN $\uparrow$ & EXAG $\downarrow$ & HAL $\downarrow$ & FAITH $\uparrow$
& ALIGN $\uparrow$ & EXAG $\downarrow$ & HAL $\downarrow$ & FAITH $\uparrow$
& ALIGN $\uparrow$ & EXAG $\downarrow$ & HAL $\downarrow$ & FAITH $\uparrow$ \\
\hline
\hline

\multirow{1}{*}{-}
& Pre-trained
& 3.67 & \textbf{1.98} & \textbf{1.64} & 4.02
& 3.33 & 2.42 & 2.04 & 3.62
& 3.44 & 2.28 & 1.92 & 3.75 \\

\hline

\multirow[c]{4}{*}{\shortstack[c]{Merging-based\\methods}}
& TA~\cite{ilharco2023editing}
& 3.59 & 2.19 & 2.01 & 3.80
& 3.27 & 2.59 & 2.33 & 3.45
& 3.37 & 2.46 & 2.23 & 3.57 \\

& TA + DGM
& \textbf{3.64} & \textbf{2.10} & \textbf{1.88} & \textbf{3.89}
& \textbf{3.44} & \textbf{2.41} & \textbf{2.10} & \textbf{3.64}
& \textbf{3.51} & \textbf{2.32} & \textbf{2.03} & \textbf{3.72} \\

\cline{2-14}

& TIES~\cite{yadav2023tiesmerging}
& 3.52 & 2.45 & 2.29 & 3.59
& 3.27 & 2.69 & 2.48 & 3.37
& 3.35 & 2.62 & 2.42 & 3.43 \\

& TIES + DGM
& \textbf{3.87} & \textbf{1.96} & \textbf{1.74} & \textbf{4.06}
& \textbf{3.50} & \textbf{2.30} & \textbf{2.07} & \textbf{3.71}
& \textbf{3.61} & \textbf{2.19} & \textbf{1.97} & \textbf{3.82} \\

\hline
\hline

\multirow[c]{2}{*}{\shortstack[c]{Direct\\adaptation}}
& $\tv$
& \textbf{4.55} & 1.83 & \textbf{1.20} & \textbf{4.51}
& \textbf{4.56} & 2.07 & \textbf{1.17} & \textbf{4.44}
& \textbf{4.56} & 2.00 & \textbf{1.18} & \textbf{4.46} \\

& Fine-tune
& 4.50 & 2.30 & 1.29 & 4.30
& 4.48 & 2.43 & 1.25 & 4.27
& 4.49 & 2.39 & 1.26 & 4.28 \\

\bottomrule
\end{tabular}
\end{table*}

\begin{table*}[t]
\centering

\begin{minipage}[t]{0.57\textwidth}
    \vspace{0pt}
    \centering

    \captionof{table}{
    \textbf{Global North to Global South generalization.}
    $\uparrow$/$\downarrow$ indicate higher/lower is better.
    }
    \label{tab:gn_gs}

    \vspace{4pt}
    \scriptsize
    \setlength{\tabcolsep}{5pt}
    \renewcommand{\arraystretch}{1.05}

    \resizebox{\linewidth}{!}{%
    \begin{tabular}{ll||ccc|c}
    \toprule
    \multirow{2}{*}{\textbf{Method}}
    & \multirow{2}{*}{\textbf{Training data}}
    & \multicolumn{4}{c}{\textbf{Global South}} \\
    & & ALIGN $\uparrow$ & EXAG $\downarrow$
    & HAL $\downarrow$ & FAITH $\uparrow$ \\
    \hline
    \hline

    Pre-trained & --
    & 3.33 & 2.42 & 2.04 & 3.62 \\
    \hline

    $\tv$ (Ours) & Synth (GS)
    & 4.56 & 2.07 & 1.17 & \textbf{4.44} \\

    CVA (Ours) & Synth (GS) + Real (GN)
    & 4.42 & \textbf{1.85} & 1.28 & 4.43 \\
    \hline

    Fine-tuned & Synth (GS)
    & 4.48 & 2.43 & 1.25 & 4.27 \\

    \bottomrule
    \end{tabular}%
    }
\end{minipage}
\hfill
\begin{minipage}[t]{0.41\textwidth}
    \vspace{0pt}
    \centering

    \captionof{table}{
    \textbf{Human preference for real vs.\ synthetic cultural vectors.}
    Pairwise preferences from native annotators across
    10 countries (\%). Best results are in \textbf{bold}.
    }
    \label{tab:human_real_synthetic}

    \vspace{4pt}
    \scriptsize
    \setlength{\tabcolsep}{5pt}
    \renewcommand{\arraystretch}{1.05}

    \resizebox{\linewidth}{!}{%
    \begin{tabular}{l||cc}
    \toprule
    \textbf{Region}
    & \textbf{Synthetic Preferred}
    & \textbf{Real Preferred} \\
    \hline
    \hline

    Global North (GN)
    & \textbf{64.9\%} & 35.1\% \\

    Global South (GS)
    & \textbf{61.0\%} & 39.0\% \\
    \hline
    \hline

    \textbf{Overall}
    & \textbf{62.1\%} & 37.9\% \\

    \bottomrule
    \end{tabular}%
    }
\end{minipage}
\vspace{-15pt}
\end{table*}

\revision{
\textbf{Composition of cultural vectors.}
Tab.~\ref{tab:tv_progressive} examines the effect of
progressively merging cultural vectors. The results show
that combining additional vectors can introduce trade-offs
between performance on different countries. For instance, we observe that adding $\tv_{iran}$ increases FAITH for Iran and Turkey while degrading performance on the USA and China. This indicates that the cross-cultural interference may be related to cultural similarities between countries. Additionally, the high similarity we observed in Fig.~\ref{fig:svd_activation} may help explain the interference among cultural vectors, which we improve by CAP.}

\revision{
\textbf{Cultural Activation Preservation (CAP).}
To address interference during composition, CAP regularizes
culture-specific fine-tuning by penalizing changes in the
pretrained model's activations on non-target cultural samples.
As discussed in Sec.~\ref{sec:result_cross_cultural_generation},
CAP improves cultural faithfulness under vector composition.
These results suggest that preserving non-target activations
can mitigate some of the interference introduced by merging,
although it does not eliminate the compositionality challenge. Additionally, we observe that CAP increases the effective rank of the stacked activation matrix (Sec.~\ref{sec:method}) across three different sources: real images (Fig.~\ref{fig:evr-cap-real-images}), private models that were used to train the model (Fig.~\ref{fig:evr_cap_private_t2i}), and public T2I models that were unseen during training (Fig.~\ref{fig:evr_cap_public_t2i})
}

\revision{Furthermore, Tab.~\ref{tab:cap_ablation} presents a four-country ablation involving Iran, the USA, China, and Turkey. Following the progressive merging setting in Tab.~\ref{tab:tv_progressive}, we progressively combine cultural vectors from Iran, the USA, and China, evaluating ALIGN across all four countries, including Turkey, whose vector is not included in the merge. Naive merging does not consistently improve ALIGN over the pretrained model, particularly when all three vectors are combined. We compare Cultural Activation Preservation (CAP) with output preservation, which regularizes the U-Net's predicted noise to remain close to that of the pretrained model on non-target cultures. CAP generally achieves higher ALIGN than output preservation across the evaluated configurations. In particular, $\lambda_{\mathrm{CAP}}=0.1$ achieves the highest scores in two of the three merging configurations, supporting the effectiveness of activation preservation in reducing cross-cultural interference. }

\textbf{Descriptor-Guided Merging (DGM) across regions.}
Tab.~\ref{tab:leave_one_out_full} provides the full results of leave-one-country-out generalization on CULTIVate, separately reporting performance for Global North (GN) and Global South (GS) countries. Naive Task Arithmetic (TA) and TIES merging generally degrade performance relative to the pretrained model, whereas Descriptor-Guided Merging (DGM) improves both methods across all metrics in both regions. In particular, TIES+DGM achieves FAITH scores of 4.06 and 3.71 for GN and GS, respectively, while also reducing exaggeration and hallucination. Across all countries, TIES+DGM achieves the highest FAITH among merging methods (3.82), exceeding the pretrained baseline (3.75), although direct cultural vector adaptation achieves higher faithfulness (4.46). These results suggest that descriptor-based weighting improves the composition of existing cultural vectors for unseen countries.

\textbf{Global North to Global South generalization.}
We evaluate whether cultural vectors can generalize to
Global South countries using only Global North real data,
simulating a setting where real data is available for
higher-resource regions but limited for underrepresented
ones. We assume access to synthetic data for all countries
but restrict real training data to Global North countries.

We compare cultural vectors learned from synthetic target
data with Cross-Cultural Vector Analogy (CVA), which adjusts
these vectors using a real-to-synthetic offset transferred
from Global North source countries.

\revision{
Tab.~\ref{tab:gn_gs} shows that cultural vectors learned from synthetic Global South data substantially improve cultural faithfulness over the pretrained model (FAITH 4.44 vs.\ 3.62). Applying Cross-Cultural Vector Analogy (CVA) further reduces cultural exaggeration (EXAG 1.85 vs.\ 2.07) while preserving comparable faithfulness (FAITH 4.43 vs.\ 4.44). These results suggest that transferring real-to-synthetic offsets across cultures can help mitigate exaggeration without compromising cultural faithfulness.
}


\subsection{Human Evaluation}
\label{app:human_eval}

To complement automatic evaluation, we conduct human studies to assess the cultural faithfulness of generated images. We recruit native annotators through Prolific, screening participants by nationality to match the evaluated countries. We randomize image presentation and include attention checks for quality control. Annotators are compensated at a rate exceeding 8 USD per hour, with approximately two annotations collected per prompt.

\revision{
\textbf{Real vs synthetic cultural vectors.}
We conduct a head-to-head comparison of cultural vectors learned from real and synthetic data across 10 countries, including three Global North countries (USA, Germany, and France) and seven Global South countries (Turkey, China, India, Nigeria, Brazil, Mexico, and Iran). Annotators select the more culturally faithful image for each prompt. As shown in Tab.~\ref{tab:human_real_synthetic}, synthetic-data vectors are preferred in 62.1\% of judgments overall, including 64.9\% in Global North and 61.0\% in Global South countries. At the prompt level, synthetic data vectors receive the majority preference for 138 of 274 prompts (50.4\%), compared with 68 prompts (24.8\%) for real data vectors, while the remaining 68 prompts (24.8\%) are tied. These results provide further human evidence supporting the effectiveness of synthetic data for learning cultural vectors.
}

\revision{
\textbf{Cross-cultural generalization.}
We evaluate Descriptor-Guided Merging (TIES+DGM) against standard TIES~\cite{yadav2023ties} and the pretrained model by asking annotators to rank images based on cultural faithfulness. The study covers 11 countries across Global North (France, Germany, Spain, USA) and Global South (Brazil, China, India, Iran, Mexico, Nigeria, Turkey), comprising 301 prompts. As shown in Tab.~\ref{tab:human_eval} (main paper), TIES+DGM achieves the lowest mean rank overall (1.72), outperforming both the pretrained model (2.08) and TIES (2.20). This improvement holds across Global North (1.66 vs. 2.08 and 2.26) and Global South (1.75 vs. 2.09 and 2.16), with TIES+DGM achieving the best mean rank in 10 of 11 countries. Pairwise Wilcoxon signed-rank tests with Holm correction confirm significant improvements over the pretrained model ($p=2.6\times10^{-7}$) and TIES ($p=5.4\times10^{-13}$). Inter-annotator agreement is also substantial (Kendall's $W=0.71$).
}





\section{Broader Impact}
\label{sec:borader_impact}

This work studies culturally grounded generation in
text-to-image models, particularly for underrepresented
cultures that may be inaccurately or stereotypically
represented in current systems. We hope our findings
contribute to more inclusive and controllable generative
models and support future research on culturally aware
generation.

\revision{
At the same time, cultural representations learned from
country-level datasets may oversimplify the diversity
within cultures or reinforce biases present in training
images and cultural descriptors. Scaling cultural vectors
may also amplify stereotypes or introduce unsupported
cultural elements. Therefore, improvements in cultural
faithfulness should not be interpreted as eliminating
bias or ensuring accurate representation of all communities
within a country.
}


\section{Evaluation Prompts}
\label{sec:prompts}
\label{app:prompts}

This section provides the prompts used to evaluate
cultural alignment, exaggeration, and hallucination.





\begin{table}
\centering
\renewcommand{\arraystretch}{1.2}
\setlength{\tabcolsep}{3pt}
\caption{ALIGN MLLM-as-a-Judge Prompt}
\label{tab:likert}
\begin{tabular}{p{0.9\linewidth}}
\hline
\textbf{ALIGN Likert Scoring Prompt} \\
\hline

\textbf{System}: You are a helpful assistant scoring how well an image depicts a prompt as it would typically appear in the specified country. Provide a rating from 1 to 5 using the guidelines below. Only output the score in the required format. \\
\\

\textbf{Scoring Scale}: \\
1 = Not well at all (completely wrong or unrelated) \\
2 = Slightly relevant \\
3 = Somewhat accurate \\
4 = Mostly accurate \\
5 = Exactly as expected for this country \\
\\

\textbf{Task}: \\
PROMPT: \texttt{\{\{prompt\}\}} \\
COUNTRY: \texttt{\{\{country\}\}} \\
\\

\textbf{User}: Respond with ONLY the score in the exact format below: \\
score is $<num>$ \\
\\

Do not include any other text, explanation, or formatting. \\
\hline
\end{tabular}
\end{table}

\begin{table}
\centering
\caption{{EXAG MLLM-as-a-Judge Prompt}}
\label{tab:EXAG_LIKERT}
\begin{tabular}{p{0.9\linewidth}}
\hline
\textbf{EXAG MLLM-as-a-Judge Prompt} \\
\hline

\textbf{System}: You are a helpful assistant scoring how exaggerated an image is in terms of cultural elements (e.g., attire, objects, background, art/design). Rate on a scale from 1 to 5 using the rules below. Only output the score in the required format. \\
\\

\textbf{Scoring Scale}: \\
1 = Appropriate and balanced. \\
2 = Slightly decorative but within cultural norms. \\
3 = Several features feel idealized or overly emphasized. \\
4 = Many elements feel exaggerated and unrealistic. \\
5 = Strongly stereotypical, misleading, or culturally irrelevant. \\
\\

\textbf{Task}: \\
PROMPT: \texttt{\{\{prompt\}\}} \\
COUNTRY: \texttt{\{\{country\}\}} \\
\\

\textbf{User}: Respond with ONLY the score in the exact format below: \\
score is $<num>$ \\
\\

Do not include any other text, explanation, or formatting. \\
\hline
\end{tabular}
\end{table}

\begin{table}
\centering
\caption{HAL MLLM-as-a-Judge prompt}
\label{tab:HAL_LIERT}
\begin{tabular}{p{0.9\linewidth}}
\hline
\textbf{HAL MLLM-as-a-Judge Prompt} \\
\hline

\textbf{System}: You are a helpful assistant scoring whether an image contains incorrect elements with respect to the mentioned activity and country. Rate on a scale from 1 to 5 using the rules below. Only output the score in the required format. \\
\\

\textbf{Scoring Scale}: \\
1 = Not at all. All elements seem appropriate. \\
2 = Slightly. One minor element seems foreign. \\
3 = Somewhat. A few elements feel unusual. \\
4 = Mostly. Many elements appear incorrect or from another culture. \\
5 = Strongly. The image is completely wrong, most elements are wrong, the scene is wrong, or belongs to another country. \\
\\

\textbf{Task}: \\
PROMPT: \texttt{\{\{prompt\}\}} \\
COUNTRY: \texttt{\{\{country\}\}} \\
\\

\textbf{User}: Respond with ONLY the score in the exact format below: \\
score is $<num>$ \\
\\

Do not include any other text, explanation, or formatting. \\
\hline
\end{tabular}
\end{table}




\end{document}